\documentclass[10pt,twocolumn,letterpaper]{article}

\usepackage[pagenumbers]{cvpr} 
\usepackage{threeparttable}
\usepackage{multirow}
\usepackage{array}

\usepackage{xcolor}
\definecolor{cvprblue}{rgb}{0.21,0.49,0.74}
\usepackage[pagebackref,breaklinks,colorlinks,allcolors=cvprblue]{hyperref}

\def\paperID{193} 
\def\confName{CVPR}
\def\confYear{2025}

\title{HuViDPO: Enhancing Video Generation through Direct Preference Optimization for Human-Centric Alignment}

\author{
Lifan Jiang$^{1}$ \quad Boxi Wu$^{1}$ \quad Jiahui Zhang$^{1}$\\
\bigskip
Xiaotong Guan$^{2}$ \quad Shuang Chen$^{1}$\\
\bigskip
$^{1}$State Key Laboratory of CAD \& CG, Zhejiang University\\
$^{2}$College of Software Technology, Zhejiang University
}

\begin{document}
\maketitle
\begin{figure*}[t]
  \centering
    \includegraphics[width=\textwidth]{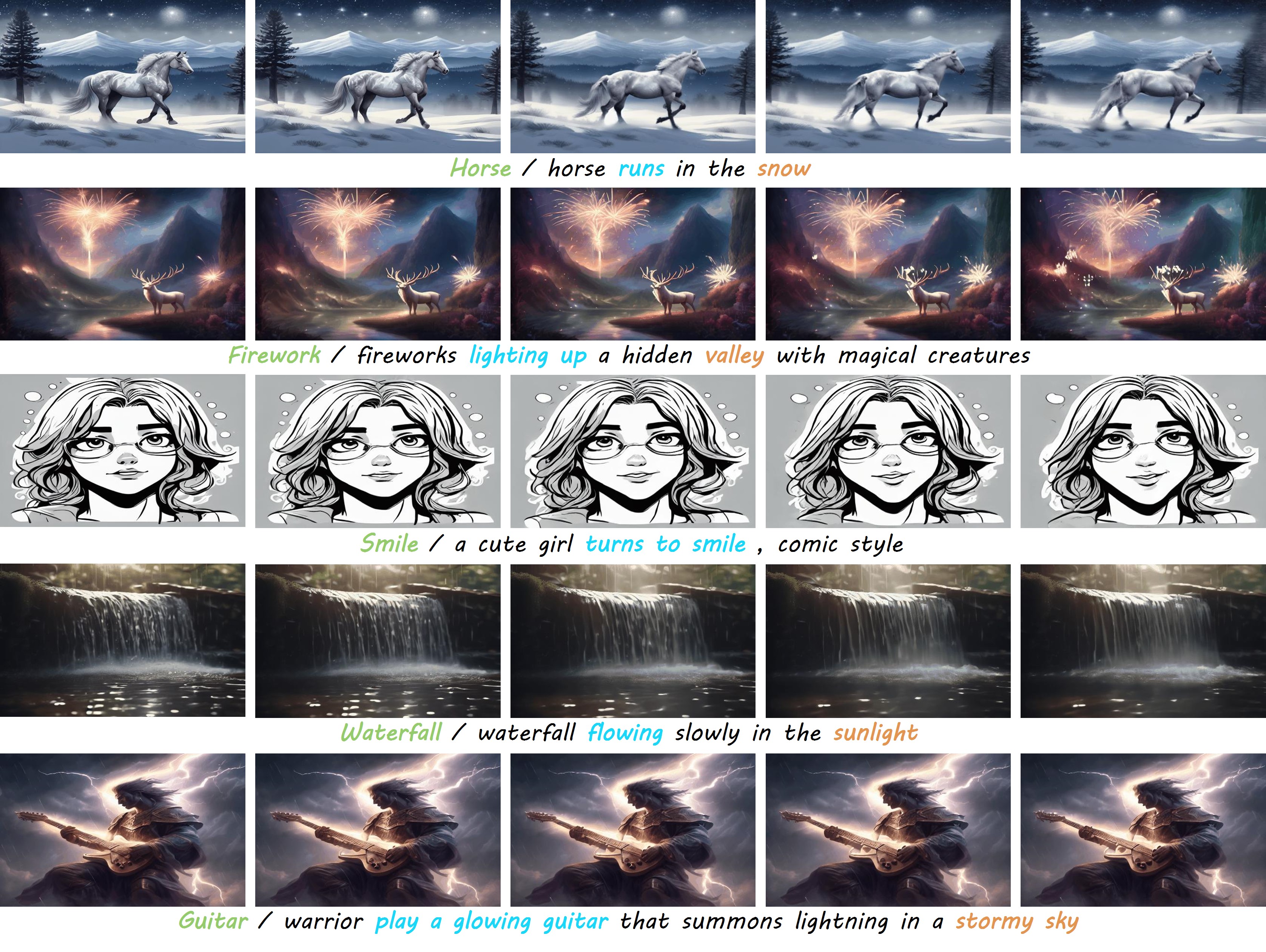}

   \caption{Videos generated by our HuViDPO. Videos generated by HuViDPO show improved flexibility, quality, and better alignment with human preferences. \textcolor[HTML]{94CB6B}{Green} highlights indicate the action category, \textcolor[HTML]{18D8FA}{blue} highlights represent motion-related information, and \textcolor[HTML]{E3954F}{orange}
highlights denote the background. You can see more examples in the supplementary material.More details and examples can be accessed on our website: \url{https://tankowa.github.io/HuViDPO.github.io/}.}
   \label{fig:1}
\end{figure*}
\begin{abstract}
With the rapid development of AIGC technology, significant progress has been made in diffusion model-based technologies for text-to-image (T2I) and text-to-video (T2V). In recent years, a few studies have introduced the strategy of Direct Preference Optimization (DPO) into T2I tasks, significantly enhancing human preferences in generated images. However, existing T2V generation methods lack a well-formed pipeline with exact loss function to guide the alignment of generated videos with human preferences using DPO strategies. Additionally, challenges such as the scarcity of paired video preference data hinder effective model training. At the same time, the lack of training datasets poses a risk of insufficient flexibility and poor video generation quality in the generated videos. Based on those problems, our work proposes three targeted solutions in sequence. 1) Our work is the first to introduce the DPO strategy into the T2V tasks. By deriving a carefully structured loss function, we utilize human feedback to align video generation with human preferences. We refer to this new method as \textbf{HuViDPO}.  
2) Our work constructs small-scale human preference datasets for each action category and fine-tune this model, improving the aesthetic quality of the generated videos while reducing training costs.  
3) We adopt a First-Frame-Conditioned strategy, leveraging the rich information from the first frame to guide the generation of subsequent frames, enhancing flexibility in video generation. At the same time, we employ a SparseCausal-Attention mechanism to enhance the quality of the generated videos.  
More details and examples can be accessed on our website: \url{https://tankowa.github.io/HuViDPO.github.io/}.
\end{abstract}
    
\section{Introduction}
\label{sec:intro}

The rise of diffusion models has significantly impacted generative models, especially in T2I generation \cite{nichol2021glide,ramesh2022hierarchical,saharia2022photorealistic,rombach2022high}. These T2I technologies have also driven the development of T2V generation \cite{he2022latent,guo2023animatediff,wu2021godiva,yu2022generating,wu2023lamp}, which aims to generate coherent video sequences from text.

Recently, there has been a surge of work that introduces Reinforcement Learning from Human Feedback (RLHF) into T2I tasks~\cite{lee2023aligning,fan2024reinforcement}, aligning the generated images with human preferences. Notably, some works \cite{wallace2024diffusion,yang2024using} have incorporated DPO strategies into T2I generation, eliminating the constraints of reward models and enhancing training efficiency.
However, current T2V generation methods face a fundamental issue: the lack of an effective and feasible loss function to guide the alignment of generated videos with human preferences during the training phase using DPO strategy. To address this issue, we rigorously derive the loss function and integrate DPO strategy into the T2V tasks for the first time. This allows us to fine-tune~\cite{an2023latent,ge2023preserve,ho2022imagen,luo2023videofusion} the model to generate videos that align with human aesthetics without the need for a separate reward model. This strategy reduces the reliance on computational resources and enhances the aesthetic quality of video generation, and we refer to it as \textbf{HuViDPO}.

Building and training a general-purpose, robust pipeline from scratch is an extremely challenging task. One of the most difficult issues is the lack of paired video preference data. Constructing a comprehensive paired video preference dataset clearly requires a substantial amount of manpower and time, but our approach cleverly circumvents this problem. For each small action category, we create corresponding small-scale human preference datasets through random pairings based on 8 to 16 human-rated videos. Subsequently, each category is fine-tuned separately using its respective dataset. Experiments show that our approach can reduce the total training duration for each category to under one day while ensuring that the model successfully learns the human preference information embedded in the small-scale human preference datasets.

The above dataset construction method, while addressing the problem, also introduces new challenges. Due to the limited data, issues such as a lack of flexibility in generated content and low video quality may arise, potentially affecting the model's generalization ability and the temporal consistency between frames. To address this, we build on the LAMP~\cite{wu2023lamp} framework and use DPO-SDXL~\cite{wallace2024diffusion} model to guide the generation of the first frame of the video, enhancing both human preference alignment and generation flexibility. Furthermore, we have improved the SparseCausal-Attention mechanism in our model, which experimental results show effectively enhances spatiotemporal consistency and improves overall video quality.

We evaluated HuViDPO using eight action categories. The experiments showed that, after simple fine-tuning, HuViDPO can generate videos corresponding to each action category and effectively transfer to different styles of video generation tasks. Compared to other baselines~\cite{he2022latent,guo2023animatediff,wu2023lamp}, the videos generated by our method align more closely with human aesthetic preferences. Some of the high-quality videos we generated are shown in \cref{fig:1}, with more of our generated videos available in the supplementary material. Our main contributions can be summarized as follows:

\begin{itemize}
\item
Integration of the DPO strategy into T2V tasks is achieved by designing a feasible loss function. The derivation process of this loss function can be found in~\cref{sec:q}. This integration enables the model to generate videos that align with human aesthetic preferences without the need for a separate reward model, thereby enhancing the aesthetic quality of the videos.

\item
Efficient DPO-Based Fine-tuning strategy using small-scale human preference datasets. This strategy not only enables efficient training of each action category within a day on a 24GB RTX4090 GPU using a small video dataset but also enhances alignment with human aesthetic preferences through DPO-Based Fine-tuning strategy, reducing costs and significantly improving preference consistency in generated videos.

\item
Enhanced video flexibility and quality with First-Frame-Conditioned strategy and SparseCausal-Attention mechanism. Leveraging DPO-SDXL for first-frame guidance aligns with human preferences and maintains flexibility, while SparseCausal-Attention mechanism enhancements improve spatiotemporal consistency and elevate video quality, creating a more coherent viewing experience.
\end{itemize}
\section{Related work}
\label{sec:formatting}

\subsection{Text-to-Video Generation}

T2V generation has made notable progress, evolving from early GAN-based models \cite{saito2017temporal,tulyakov2018mocogan,fu2023tell,li2018video,wu2022nuwa,yu2022generating} to newer transformer \cite{yan2021videogpt,arnab2021vivit,esser2021taming,ramesh2021zero,yu2022scaling} and diffusion models \cite{kirkpatrick2017overcoming,sohl2015deep,song2020denoising,zhang2022gddim}. Early efforts like MoCoGAN~\cite{tulyakov2018mocogan} focused on short video clips but faced issues with stability and coherence. The introduction of transformers improved sequential data handling, enhancing video generation, while diffusion models further improved video quality by progressively denoising the input. 
Despite these advances, T2V models still struggle to reflect human preferences, with the generated videos generally lacking aesthetic quality. Additionally, the scarcity of paired video preference data hinders effective model training and may lead to insufficient flexibility and poor quality in the generated videos.

\subsection{RLHF}

RLHF \cite{gao2023scaling,stiennon2020learning,rafailov2024direct} is a method that utilizes human feedback to guide machine learning models. Early RLHF algorithms, such as DDPG~\cite{lillicrap2015continuous} and PPO~\cite{schulman2017proximal}, typically relied on complex reward models to quantify human feedback. These reward models require a large amount of annotated data and face challenges during tuning. As research has progressed, more efficient preference learning methods have emerged, among which DPO has become a new framework. DPO does not depend on a separate reward model; instead, it obtains human preferences through pairwise comparisons and directly optimizes these preferences. This shift not only simplifies the application of RLHF but also enhances the alignment of models with human values. Furthermore, DPO has been successfully introduced into T2I tasks~\cite{wallace2024diffusion,yang2024using}, providing new insights for generative models in addressing the alignment of human preferences and showcasing DPO's potential in the field of AIGC~\cite{shi2024instantbooth,
qing2024hierarchical,menapace2024snap,koley2024s}. However, there remains a gap in current research regarding the application of DPO strategies to T2V tasks. Effectively integrating DPO into T2V tasks presents a challenging endeavor.

\section{Method}
\label{sec:2}

In this section, we first introduce the derivation of the loss function, which is used to apply the DPO strategy to T2V tasks in~\cref{sec:q}. Next, we provide a detailed description of the DPO-Based Fine-tuning strategy by using small-scale human preference datasets in~\cref{sec:w}. Then, we detail the First-Frame-Conditioned strategy and the proposed SparseCausal-Attention mechanism in~\cref{sec:e}. Finally, in~\cref{sec:r}, we introduce the inference process of HuViDPO and its powerful generation capabilities.

\begin{figure*}[t]
  \centering
    \includegraphics[width=\textwidth,height=0.4\textheight]{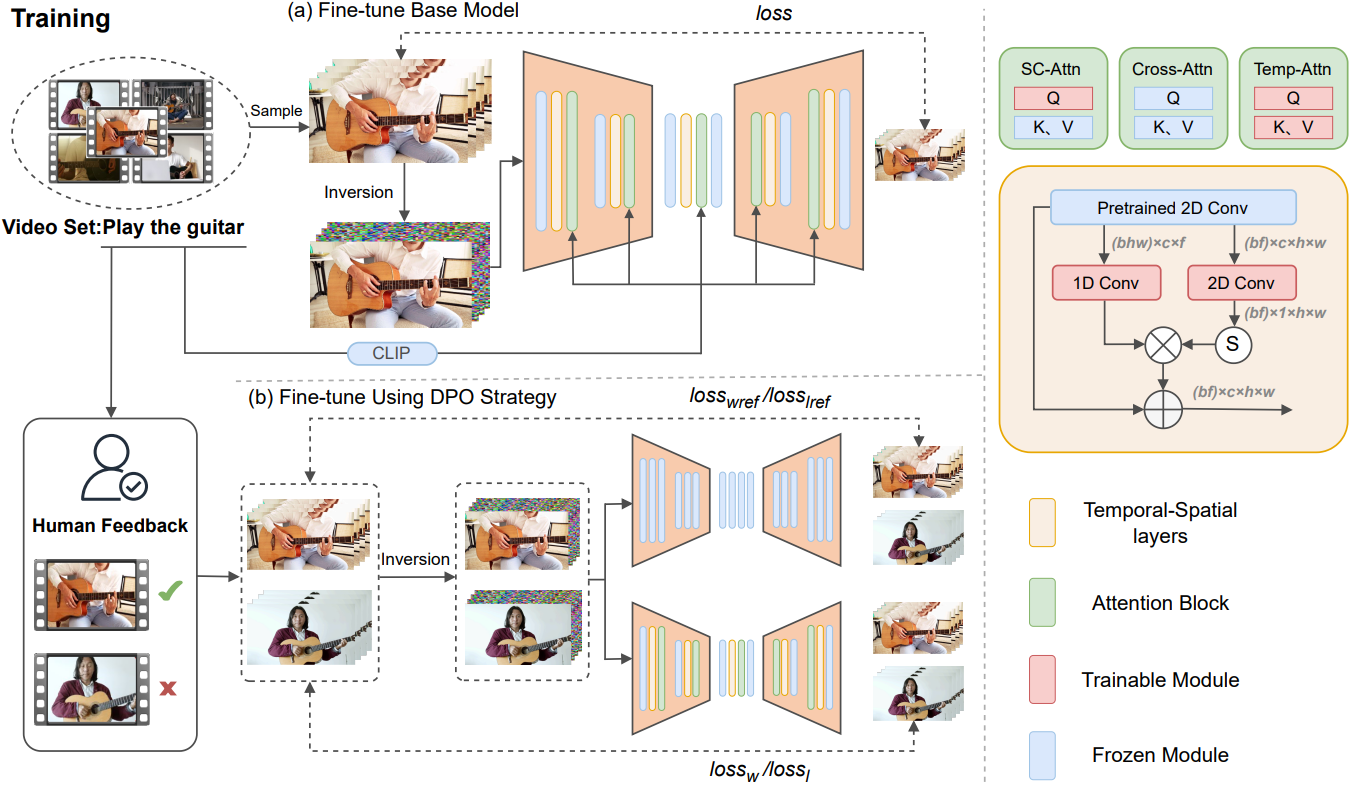}

   \caption{Training pipeline of our HuViDPO. Training process can be divided into two stages: (a) Training the Attention Block and Temporal-Spatial layers using basic training settings to improve the spatiotemporal consistency. (b) Fine-tuning the model, with LoRA added and other layers frozen, using small-scale human preference datasets and DPO strategy to enhance its alignment with human preferences. In phase (b), $loss_w$ and $loss_l$ denote the loss values computed by inputting winning and losing videos into the fine-tuned model, while $loss_{wref}$ and $loss_{lref}$ are the loss values obtained by inputting the same videos into the reference model.}
   \label{fig:2}
\end{figure*}

\subsection{Integrating DPO into T2V Tasks via Loss Function Design}
\label{sec:q}

In this section, we provide a detailed derivation of our unique loss function. We extend the DPO strategy's loss function from image generation to video generation. The detailed derivation of this strategy for T2I tasks is provided in the appendix, and the loss is defined as:

\begin{equation}
\label{eq:1}
\mathcal{L}_{\text{Image}}(\theta) = -\mathbb{E}_{(x_{0}^w, x_{0}^l) \sim \mathcal{D}} \log \sigma \left( \beta \left( \Delta(x_{0:T}^w) - \Delta(x_{0:T}^l) \right) \right),
\end{equation}

where \(x_0^w\) and \(x_0^l\) are the winning and losing image samples, \(\beta\) is a hyperparameter, and  indicates the corresponding time step, and \(\Delta(x)\) is defined as:

\begin{equation}
\label{eq}
\Delta(x) = \log \frac{p_\theta(x)}{p_{\text{ref}}(x)}.
\end{equation}

Based on~\cref{eq:1,eq}, the DPO loss function for the T2V task, where \( v_0^w \) and \( v_0^l \) are the winning and losing video samples, is defined as:

\begin{equation}
\label{eq:3}
\mathcal{L}_{\text{Video}}(\theta) = -\mathbb{E}_{(v_0^w, v_0^l) \sim \mathcal{D}} \log \sigma \left( \beta \left( \Delta(v_{0:T}^w) - \Delta(v_{0:T}^l) \right) \right).
\end{equation}

To handle the complexity of calculating high-dimensional video sequence probabilities with a total of \( T = 1000 \) time steps, we employ an approximation approach. We introduce an approximate posterior \( q(v_{1:T} | v_0) \) for the subsequent time steps and utilize the Evidence Lower Bound (ELBO)~\cite{kingma2013auto,hoffman2013stochastic} to approximate $\log p_\theta(v_{0:T})$. Then, by expressing \( p_\theta(v_{0:T}) \) and \( q(v_{1:T} | v_0) \) as products of conditional probabilities at each time step, we achieve a stepwise sampling approach. The final approximate expression is:

\begin{equation}
\label{KL1}
\begin{split}
\log p_\theta(v_{0:T}) \approx \mathbb{E}_{q(v_t | v_{t-1}), t \sim \{1..T\}} 
\Bigg[ \log \frac{p_\theta(v_0)}{q(v_0)} \\
+ \log \frac{p_\theta(v_t | v_{t-1})}{q(v_t | v_{t-1})} \Bigg].
\end{split}
\end{equation}

Since \( q(v_t | v_{t-1}) \) is a conditional probability distribution that generally sums to 1, the KL divergence~\cite{kingma2013auto,blei2017variational} can be expressed as:

\begin{equation}
\label{KL2}
\mathbb{D}_{KL}\left( q(v_t | v_{t-1}) \, \| \, p_\theta(v_t | v_{t-1}) \right) =  \log \frac{q(v_t | v_{t-1})}{p_\theta(v_t | v_{t-1})}.
\end{equation}

Based on~\cref{KL1,KL2}, we rewrite \( \log p_\theta(v_{0:T}) \) as:

\begin{equation}
\label{eq:6}
\begin{split}
\log p_\theta(v_{0:T}) \approx \mathbb{E}_{q(v_{1:T} | v_0), t \sim \{1..T\}} \left[ \log \frac{p_\theta(v_0)}{q(v_0)} \right] \\ 
- \mathbb{D}_{KL}\left( q(v_t | v_{t-1}) \, \| \, p_\theta(v_t | v_{t-1}) \right).
\end{split}
\end{equation}

Moreover, the derivation of \( \log p_{\text{ref}}(v_{0:T}) \) is consistent with that of \( \log p_\theta(v_{0:T}) \). Based o~\cref{eq,eq:6}, we can rewrite \( \Delta(v_{0:T}) \) as

\begin{equation}
\label{eq:7}
\Delta(v_{0:T}) = -\mathbb{D}_{KL}^{\theta} + \mathbb{D}_{KL}^{\text{ref}} + C.
\end{equation}

By rewriting the KL divergence in terms of noise prediction, we can express it as follows:

\begin{equation}
\label{eq:9}
\mathbb{D}_{KL}^{\theta} \propto ||\epsilon - \epsilon_\theta(v_t, t)||^2 
\quad \mathbb{D}_{KL}^{\text{ref}} \propto ||\epsilon - \epsilon_\text{ref}(v_t, t)||^2.
\end{equation}


Finally, based on~\cref{eq:3,eq:7,eq:9}, the complete form of the DPO loss function for T2V task is:

\begin{equation}
\label{lab:2}
\begin{aligned}
\mathcal{L}_{\text{Video}}(\theta) = & \, \mathbb{E}_{(v_{0}^w, v_{0}^l) \sim \mathcal{D}, \, t \sim \{1..T\}} \Bigg[\beta \log \sigma \Bigg( \\
& \quad \left( ||\epsilon_w - \epsilon_\theta(v^w_t, t)||^2 - ||\epsilon_w - \epsilon_{\text{ref}}(v^w_t, t)||^2 \right) \\
& \quad - \left( ||\epsilon_l - \epsilon_\theta(v^l_t, t)||^2 - ||\epsilon_l - \epsilon_{\text{ref}}(v^l_t, t)||^2 \right) \Bigg) \Bigg].
\end{aligned}
\end{equation}

\subsection{DPO-Based Fine-tuning strategy}
\label{sec:w}

Our training process can be roughly divided into two steps, as shown in~\cref{fig:2}. First, within the standard VDM architecture, we train using a small video dataset and fine-tuning the Attention Block and Temporal-Spatial layers to ensure spatiotemporal consistency in the generated videos. It is important to note that during training, the first frame is not subjected to noise injection, while the other frames are generated through DDIM inversion~\cite{song2020denoising} from the original training data. Seen as~\cref{fig:7}, since the first frame contains most of the key information in the short video, it plays a crucial role in guiding the generation of subsequent frames, ensuring diversity in video generation, while our method enhances the visual aesthetics of the generated images.

Next, we conduct human preference scoring on videos used for DPO strategy training to evaluate each video against aesthetic standards. After scoring, the dataset is shuffled, noted as \( V_{\text{shuffle}} = \text{shuffle}(V) \), and a random pair \( (V_a, V_b) \) is selected as a training sample. The loss function, defined as~\cref{lab:2}, incorporates the DPO strategy into the VDM loss to align video generation with human preferences. During training, a reference model with frozen weights and a training model with an added LoRA module are used. Both models generate video sequences guided by the same prompt and first-frame image, and the loss from~\cref{lab:2} updates the LoRA weights through backpropagation, enhancing video alignment with human preferences and improving quality.

It is worth noting that, due to our DPO-Based Fine-tuning strategy, training time and computational resources are greatly reduced, with each action category requiring less than a day for the above two training processes.

\subsection{Enhancing Video Flexibility and Quality}
\label{sec:e}

\begin{figure}[!h]
  \centering
    \includegraphics[width=\linewidth]{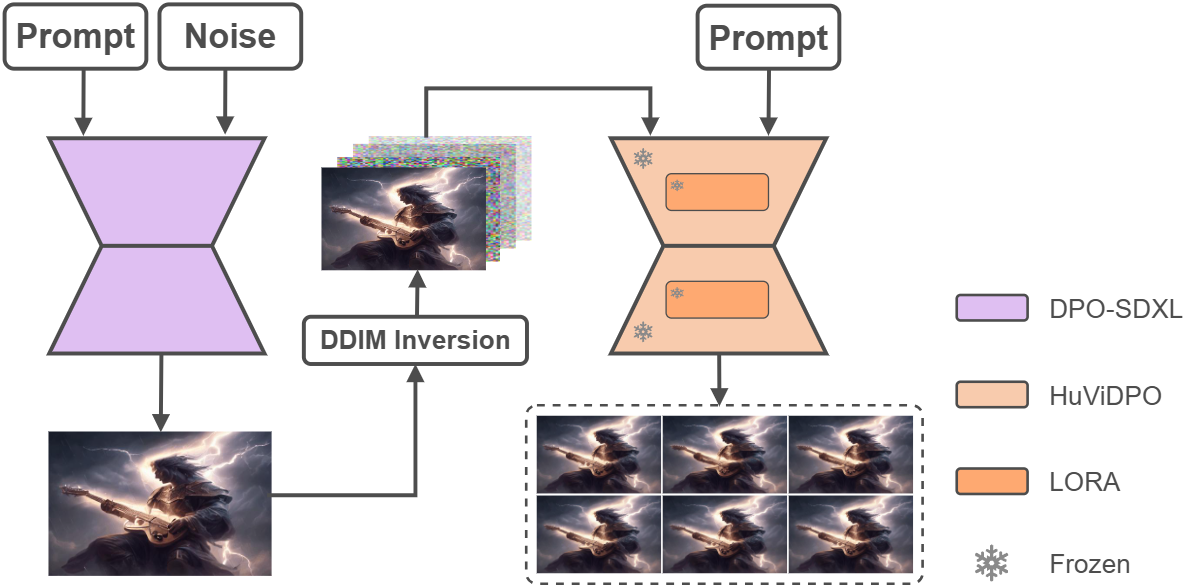}

   \caption{Inference Process of our HuViDPO. We first employ DPO-SDXL to create a diverse style-first frame, then concatenate it with other noise frames, which are then inputted into a trained model to produce the video output.}
   \label{fig:4}
\end{figure}

\noindent\textbf{First-Frame-Conditioned Generation using DPO-SDXL.}
Due to the limited paired human preference video training data, the model risks losing flexibility in video generation and lacking rich semantic expression. To address this issue, we decouple content and motion. From our general understanding, the first frame of a video contains the majority of the content information of a short video. Therefore, guiding the generation of a short video with a high-quality first frame image that contains rich semantic information can effectively address the problems of losing flexibility and lacking rich semantic expression in video generation. In our method setup, we use DPO-SDXL~\cite{wallace2024diffusion} to generate the first frame of the video to guide the generation of subsequent video frames, as shown in~\cref{fig:4}. This method not only allows for the generation of videos based on prompts of various styles, ensuring flexibility in generation, but also aligns the video production more closely with human aesthetic standards, thereby achieving an optimal balance between maintaining flexibility and the lack of datasets.

\begin{figure}[t]
  \centering
    \includegraphics[width=\linewidth]{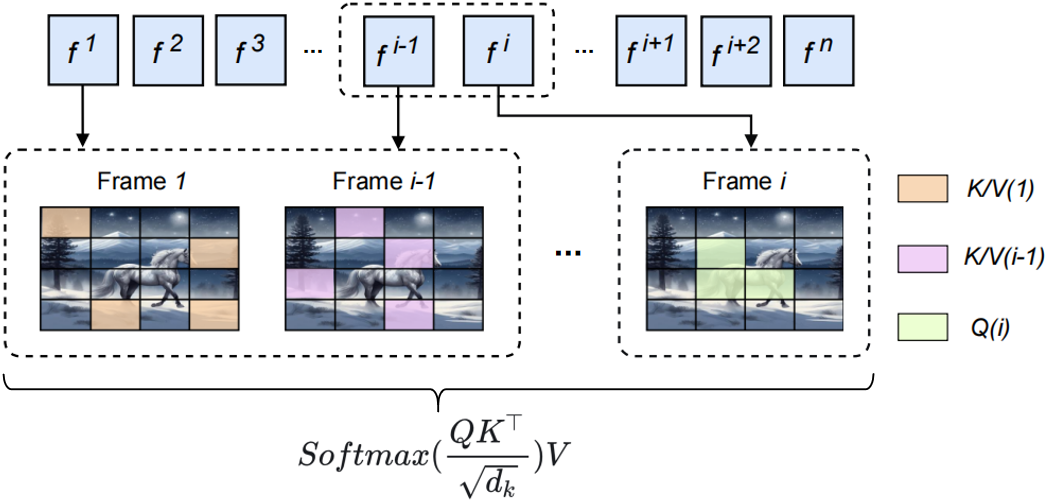}

   \caption{The details of the proposed SparseCausal-Attention Mechanism. We extract $K/V$ tokens from the first frame and the $i-1$ frame, and compute the attention mechanism with the $Q$ of the $i$ frame.}
   \label{fig:3}
\end{figure}

\vspace{0.1cm}
\noindent\textbf{Improved SparseCausal-Attention Mechanism.}
To further address the spatiotemporal inconsistency issue in the few-shot T2V generation tasks and to improve the overall quality of video generation, our method designs and enhances the SparseCausal-Attention module. In our approach, for each frame $i$ in the video, we extract key/value \((K/V)\) information from both the first frame and the previous frame $i-1$, using the query \((Q)\) information of the current frame $i$ to compute the required attention mechanism~\cite{xing2024simda,wang2023crossformer++}. The attention calculation is formulated as follows:

\begin{equation}
\text{Attention}(Q_i, K, V) = \text{Softmax}\left(\frac{Q_i K_{\text{concat}}^\top}{\sqrt{d_k}}\right) V_{\text{concat}},
\label{eq:attention_calculation}
\end{equation}

where \(K_{\text{concat}}\) and \(V_{\text{concat}}\) represent the concatenated keys and values, defined as:

\begin{equation}
K_{\text{concat}} = [K_0; K_{i-1}] \quad V_{\text{concat}} = [V_0; V_{i-1}].
\label{eq:concat_key}
\end{equation}

As shown in~\cref{fig:8}, compared to methods without the improved SparseCausal-Attention module, this enhancement maintains frame consistency in the generated videos under equivalent conditions, leading to outputs that better align with human aesthetic standards and improve video quality. 

\subsection{Inference Process of HuViDPO}
\label{sec:r}

Similar to the training process, the inference process can also be divided into two steps. As shown in~\cref{fig:4}, we first use DPO-SDXL to generate the first-frame image based on the corresponding text prompt. This image aligns better with human preferences and provides a richer diversity of styles. Next, the generated first-frame image is combined with other frames composed of Gaussian noise to form a video sequence. This sequence, along with the same text prompt, is then input into the model that has been fine-tuned twice as described in~\cref{sec:w}, to generate the final video output. The cases presented in this paper primarily come from eight different action categories and various video styles. ~\cref{fig:1} and the appendix provide an initial demonstration of the high quality and strong alignment with human aesthetics of videos generated by HuViDPO. Of course, more examples are available in the supplementary material we provided, allowing readers to intuitively experience the video quality. It is worth noting that HuViDPO offers a high degree of generation flexibility, allowing users to customize different action categories and randomly select desired video styles, making it highly practical.

\begin{figure*}[t]
  \centering
    \includegraphics[width=\textwidth,height=188pt]{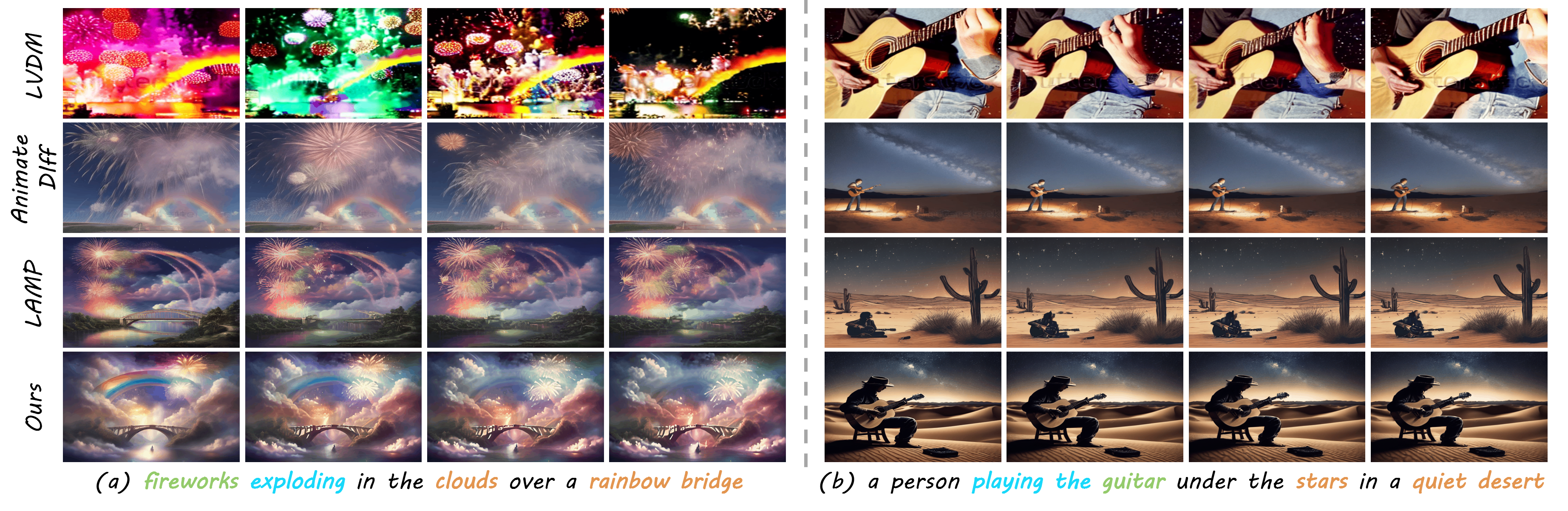}

   \caption{Qualitative comparison with LVDM~\cite{he2022latent}, AnimateDiff~\cite{guo2023animatediff}, and LAMP~\cite{wu2023lamp}. The above images clearly demonstrate that the videos generated by our method exhibit better spatiotemporal consistency and are more visually aligned with human preferences. In example (a) of this image, video generated by our method exhibits richer content and stronger spatiotemporal consistency, such as the rainbow and the shape of the bridge remaining largely unchanged. In example (b) of this image, video generated by our method features a more reasonable layout and richer main subjects, such as more detailed character features and a softer background.} 
   \label{fig:5}
\end{figure*}

\section{Experiment}

\subsection{Implementation Detail}

Our training was conducted on a single 24G RTX4090 GPU across eight small-scale human preference datasets, including categories like waterfall, smile and guitar. We fine-tuned the SparseCausal-Attention module, Temporal-Attention module and Temporal-Spatial layers with a learning rate of 3e-5, batch size of 1, and training iters of 25000. Based on~\cref{lab:2}, the loss was calculated with a learning rate of 1e-4, along with weight decay and epsilon adjustments to prevent gradient issues. 

In comparing with baselines, we use standard evaluation metrics such as CLIPscore~\cite{radford2021learning}, SSIM~\cite{wang2004image}, and MSE, benchmarking our results against LVDM~\cite{he2022latent}, AnimateDiff~\cite{guo2023animatediff}, and LAMP~\cite{wu2023lamp}. To further highlight our approach's superiority, we visually present the quality of videos generated by different methods under identical prompts, allowing a direct comparison that underscores our method’s advantages. Additionally, we conducted a user study comparing our approach with baselines to showcase its effectiveness. Finally, an ablation study tested the importance of each module designed in ~\cref{sec:2}, reinforcing the scientific rigor of our approach.


\begin{table}[]
\centering
    {
    \setlength{\extrarowheight}{2pt}
    \scalebox{0.72}{
        \begin{threeparttable}
        \begin{tabular}{@{}l@{\hskip 4pt}l@{\hskip 2pt}c@{\hskip 2pt}c@{\hskip 2pt}c@{\hskip 2pt}c@{\hskip 3pt}c@{\hskip 3pt}c@{\hskip 3pt}c@{}}
            \toprule
                Metric                & Method                  & waterfall            & smile         & helicopter      & rain          & firework      & guitar      & All        \\ \midrule
                \multirow{4}{*}{CLIPscore $\mathrel{\uparrow}$} 
                & LVDM~\cite{he2022latent}              & 31.1         & 26.1          & 31.8       & 30.4            & 29.3          & 33.0       & 30.3          \\
                & AnimateDiff~\cite{guo2023animatediff}           & \underline{34.1}          & 27.7          & 29.4        & 31.3            & 31.8            & 31.8        & 31.0          \\
                & LAMP~\cite{wu2023lamp}             & 33.6          & \textbf{28.9}          & \underline{33.0}        & \textbf{33.2}            & \textbf{34.0}         & \underline{35.8}        & \textbf{33.1}          \\
                & \textbf{Ours}    & \textbf{34.9}          & \underline{28.1}       & \textbf{33.1}            & \underline{31.9}            & \underline{32.5}        & \textbf{37.2}  & \underline{33.0}        \\ \midrule

                \multirow{4}{*}{SSIM $\mathrel{\uparrow}$} 
                & LVDM~\cite{he2022latent}     & 73.1          & 54.4          & 52.0          & 67.8            & 55.1          & 65.4          & 61.3            \\
                & AnimateDiff~\cite{guo2023animatediff}     & 82.1          & 84.8          & 71.2          & 82.0          & 65.7          & 87.7          & 78.9          \\
                & LAMP~\cite{wu2023lamp}    & \textbf{95.2}          & \underline{88.1}          & \underline{81.3}          & \textbf{93.3}          & \textbf{88.0}          & \underline{93.9}            & \underline{89.9}            \\
                & \textbf{Ours} & \underline{89.3} & \textbf{92.2} & \textbf{91.6} & \underline{91.0} &\underline{86.6} & \textbf{95.8} & \textbf{91.1} \\ \midrule

                \multirow{4}{*}{MSE $\mathrel{\downarrow}$} 
                & LVDM~\cite{he2022latent}     & 62.0          & 68.0          & 66.6            & 62.7          & 82.0          & 61.3          & 67.1          \\
                & AnimateDiff~\cite{guo2023animatediff}     & 33.8          & 32.2          & 51.6          & 32.5          & 52.5          & 20.1          & 37.1          \\
                & LAMP~\cite{wu2023lamp}    & \textbf{12.2}          & \underline{25.1}            & \underline{31.5}          & \underline{12.1}          & \underline{25.9}          & \underline{14.4}            & \underline{20.2}          \\
                & \textbf{Ours} & \underline{24.3} & \textbf{17.1} & \textbf{24.1} & \textbf{8.1} & \textbf{24.8} &  \textbf{13.1} & \textbf{18.6} \\

            \bottomrule
        \end{tabular}
        
        \end{threeparttable}
        }
    }
\caption{Quantitative comparisons with baselines. Our method surpasses all baselines in SSIM and MSE, demonstrating superior inter-frame consistency. Bold font indicates the best performance, while underlined font indicates the second best.}
\label{tab:4}
\end{table}

\begin{table}
  \centering
  \setlength{\extrarowheight}{1pt}
  \scalebox{0.85}{
  \begin{tabular}{@{}lccc@{}}
    \toprule
    Method &  Aesthetics $\mathrel{\uparrow}$ & Consistency $\mathrel{\uparrow}$ & Alignment $\mathrel{\uparrow}$ \\
    \midrule
    LVDM~\cite{he2022latent} & 45.4& 48.3& 55.8\\
    AnimatedDiff~\cite{guo2023animatediff} & 67.6& \underline{61.1}& 72.4\\
    LAMP~\cite{wu2023lamp} & \underline{72.5}& 60.8& \underline{75.6}\\
    \textbf{Ours} & \textbf{83.9}& \textbf{75.1}& \textbf{85.1}\\
    \bottomrule
  \end{tabular}
  
  }
  \caption{User preference comparison. This table demonstrates our significant superiority in aspects such as Aesthetics, Consistency, and Alignment, in comparison to LVDM~\cite{he2022latent}, AnimateDiff~\cite{guo2023animatediff}, and LAMP~\cite{wu2023lamp}. Bold font indicates the best performance, while underlined font indicates the second best.}
  \label{tab:example2}
\end{table}

\begin{figure*}[t]

  \centering
  \includegraphics[width=\textwidth]{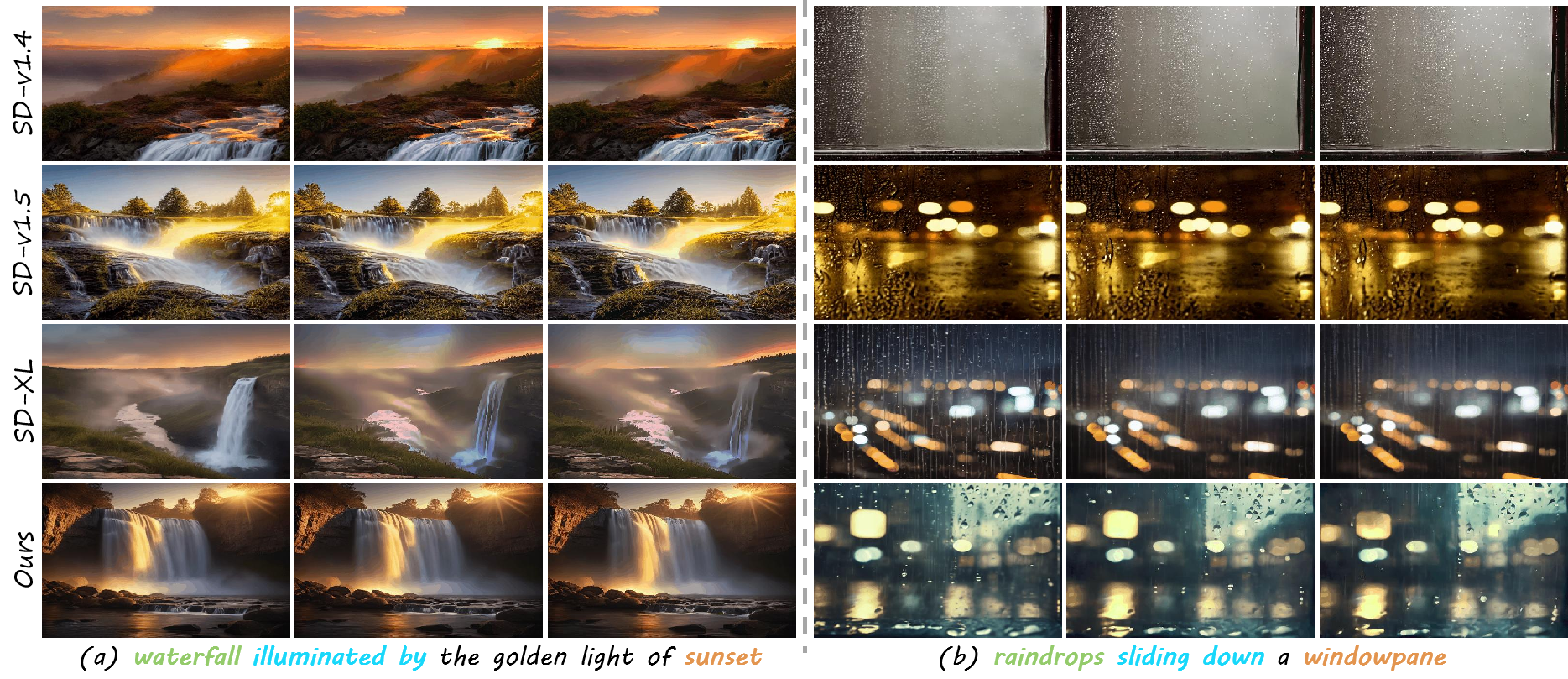}

   \caption{Qualitative comparison with SD-v1.4~\cite{rombach2022high}, SD-v1.5 and SD-XL~\cite{podell2023sdxl}. The videos generated by our method offer richer visual effects and stronger text-to-video alignment. In the left image, the generated video aligns better with the prompt, and the waterfall exhibits enhanced fluidity and structural consistency. In the right image, the video appears more realistic, natural, and smooth.}
   \label{fig:7}
\end{figure*}

\subsection{Comparison with Baseline}

\noindent\textbf{Quantitative Result.} In this experiment, We set up 10 identical prompts for each category, generating 60 videos per baseline for evaluation. CLIPscore was used to measure alignment with prompts, where higher values indicate better alignment. We also calculated SSIM and MSE for spatiotemporal consistency, with higher SSIM and lower MSE scores preferred. As shown in~\cref{tab:4}, our method outperforms LVDM and AnimateDiff on CLIPscore, indicating strong text alignment. Additionally, HuViDPO achieved higher SSIM (1.2\% above LAMP, 29.8\% and 12.2\% over LVDM and AnimateDiff) and lower MSE (1.6\% below LAMP, 48.5\% and 18.5\% below LVDM and AnimateDiff), confirming its robust spatiotemporal consistency.

\begin{figure}[!h]
  \centering
    \includegraphics[width=\linewidth, height=150px]{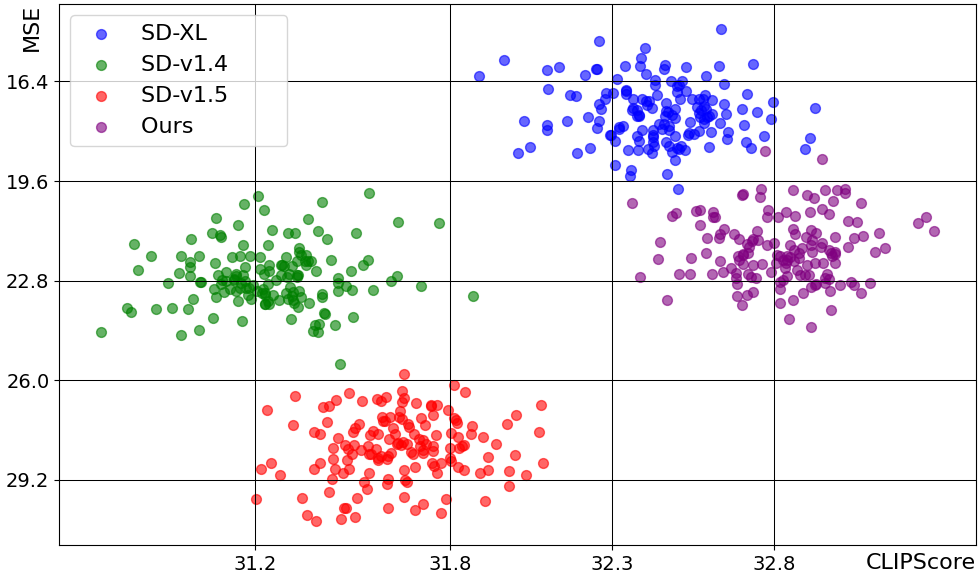}

   \caption{Comparison between our First-Frame Generation strategy and the SD-v1.4~\cite{rombach2022high}, SD-v1.5, and SD-XL~\cite{podell2023sdxl} methods. HuViDPO outperforms SD-v1.4~\cite{rombach2022high} and SD-v1.5 on both metrics, but scores slightly lower on MSE compared to SD-XL~\cite{podell2023sdxl}. The slight discrepancy in MSE is primarily due to the larger motion amplitudes in the videos, but this does not affect the overall visual effectiveness of the videos.}
   \label{fig:6}
\end{figure}

\begin{figure}[!h]
  \centering
  \includegraphics[width=\linewidth]{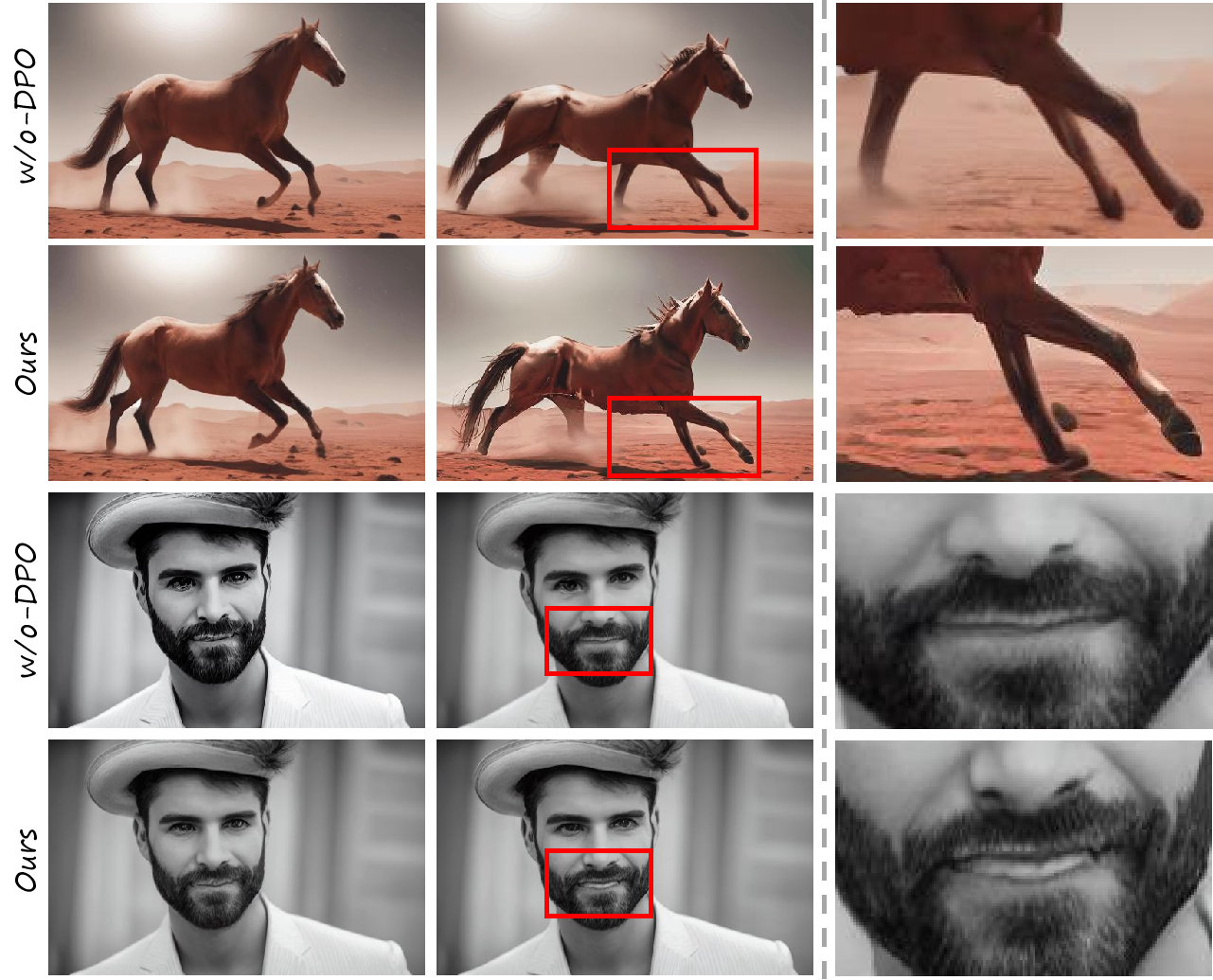}
  \caption{Qualitative comparison with methods without DPO Fine-tune. Details marked by the red box are magnified and displayed on the right side of the image. The above example shows that without DPO fine-tuning, the horse's legs appear significantly deformed. The below example demonstrates that with DPO fine-tuning, the man's smile process is more natural, revealing details of his teeth compared to without DPO fine-tuning.}
  \label{fig:9}
\end{figure}

\noindent\textbf{Qualitative Result.} First, we used a unified prompt input across different methods to generate videos, with example outputs shown in~\cref{fig:5}. These outputs clearly demonstrate that videos generated by HuViDPO exhibit greater flexibility, higher visual quality, and align more closely with human aesthetic preferences. Next, to address the limitations of existing evaluation metrics and further validate that videos generated by HuViDPO better meet human aesthetic standards, we conducted a large-scale user study. In total, 58 participants, including both experts and non-experts, evaluated videos from four different methods across 14 diverse cases. The participants rated each video on aesthetics, spatiotemporal consistency, and text alignment from 1 to 5, with the final aggregated scores converted to percentages (see Appendix for details on experimental setup). As shown in~\cref{table:2}, our method outperformed all baselines across every criterion, strongly indicating that the video animations generated by our approach better align with human preferences.

\begin{figure}[!h]
  \centering
  
  \includegraphics[width=\linewidth]{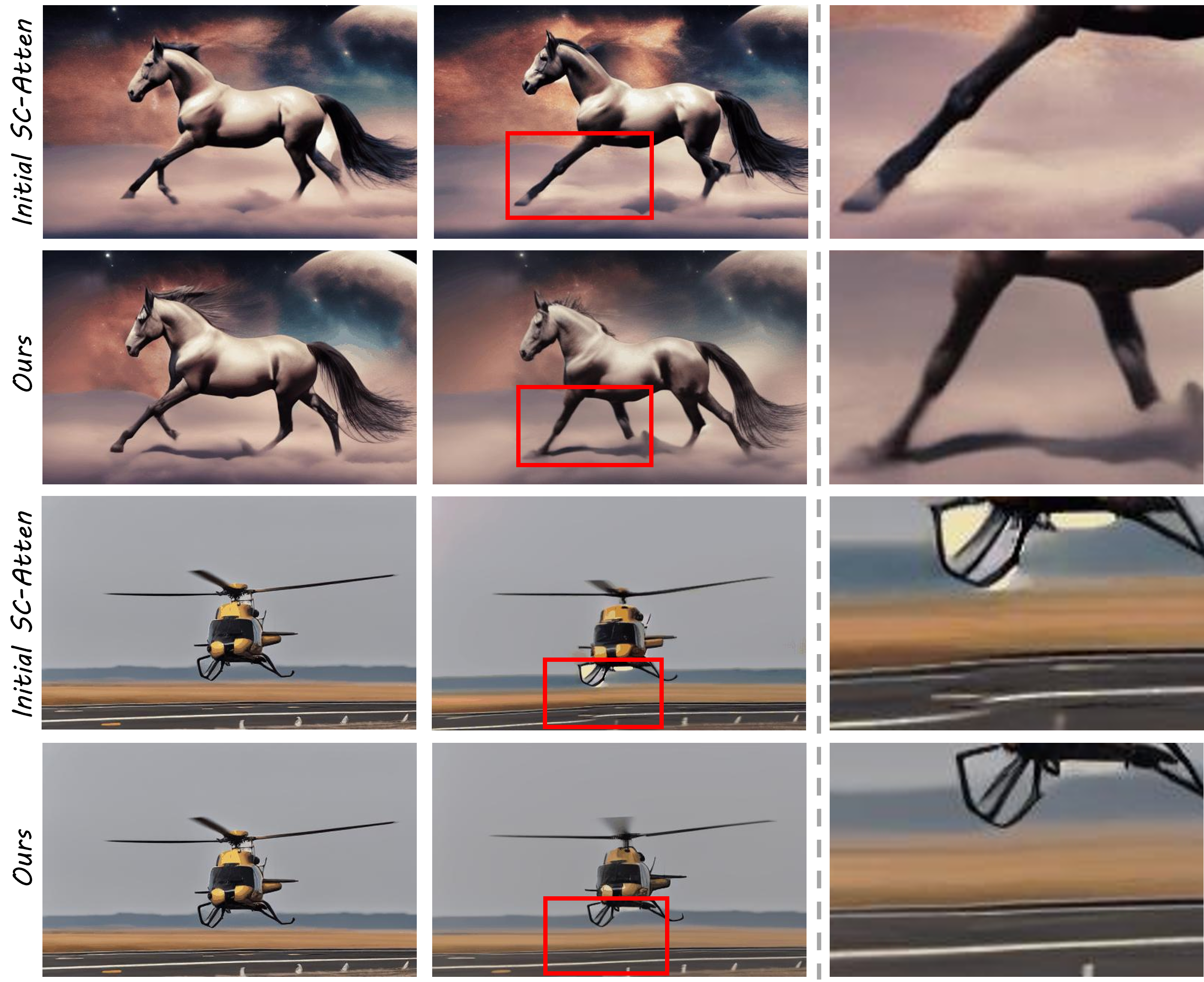}
  \caption{Qualitative comparison with unmodified SparseCausal-Attention. Red boxed details are shown magnified on the right. In the upper example, unmodified SparseCausal-Attention makes the horse's legs vanish, while our method preserves natural leg motion. In the lower example, the same unmodified mechanism distorts the road surface, ruining structural coherence, whereas our method keeps the road smooth.}
  \label{fig:8}
\end{figure}

\subsection{Ablation Study}

\noindent\textbf{Effect of the DPO-Based Fine-tuning Strategy.}
DPO-Based Fine-tuning is a crucial part of our overall strategy. Our method significantly reduces abrupt transitions between different video frames across most video categories. As shown in~\cref{fig:9}, this experiment selected two categories, smile and horse running, to visually illustrate the advantages of DPO-Based Fine-tuning. This improvement enhances the fluidity and aesthetic quality of the videos, providing a more cohesive viewing experience. Additionally, the fine-tuning process effectively aligns the generated content with human preferences, making the video transitions smoother and visually more appealing. This alignment not only enhances the visual experience but also ensures that the content resonates more strongly with viewers.

\noindent\textbf{Effect of the First-Frame-Conditioned Strategy.}
The first frame of a video contains essential semantic and content information that effectively guides the generation of subsequent frames, playing a significant role in video generation tasks. By using DPO-SDXL for first-frame generation, we ensure that the resulting videos exhibit greater flexibility and diversity. The superiority of our method is evident in~\cref{fig:6}. Notably, while our approach scores slightly worse than SD-XL on the MSE metric, this is primarily due to the larger motion amplitudes in videos generated by our method, which means this metric does not fully capture the aesthetic quality. To further substantiate and provide an intuitive assessment of the effectiveness of our First-Frame Generation strategy, we selected two specific examples in~\cref{fig:7} to visually showcase the differences between SD-v1.4, SD-v1.5, the SD-XL method, and our first-frame generation strategy. The results indicate that our strategy outperforms the others in visual quality. Additionally, to further demonstrate the flexibility, diversity, and adaptability of our generated videos, we invite you to explore more examples in the supplementary material for a comprehensive understanding.

\vspace{0.05cm}
\noindent\textbf{Effect of the Improved SparseCausal-Attention Mechanism.} 
SparseCausal-Attention is a key module in our model, effectively enhancing the spatiotemporal consistency and quality of the generated videos. This module is crucial for maintaining continuity between frames. As shown in~\cref{fig:8}, compared to the SparseCausal-Attention in the base model LAMP, our improved module further strengthens information retention between frames. This ensures smoother transitions that align better with human aesthetic preferences. As a result, the generated videos are not only more visually appealing but also exhibit higher quality.

\section{Conclusion}

In this study, we are the first to introduce DPO strategies into T2V tasks. By designing a comprehensive loss function, we utilize human feedback to align video generation with human aesthetic preferences, naming it \textbf{HuViDPO}. Additionally, we constructed small-scale human preference datasets for each action category and employed LoRA to fine-tune the model, significantly improving the aesthetic quality of the generated videos while reducing training costs. Regarding resource efficiency, our setup allows each fine-tuning process for an action category to complete in under a day on a single 24GB RTX4090 GPU. Finally, to enhance the quality and flexibility of the generated videos, we adopted a First-Frame-Conditioned strategy, utilizing the rich information provided by the first frame to guide the generation of subsequent frames, and integrated a SparseCausal-Attention mechanism to further improve video quality and spatiotemporal consistency, ensuring a smoother and more coherent video generation process.
{
    \small
    \bibliographystyle{ieeenat_fullname}
    \bibliography{main}

\begin{thebibliography}{48}
\providecommand{\natexlab}[1]{#1}
\providecommand{\url}[1]{\texttt{#1}}
\expandafter\ifx\csname urlstyle\endcsname\relax
  \providecommand{\doi}[1]{doi: #1}\else
  \providecommand{\doi}{doi: \begingroup \urlstyle{rm}\Url}\fi

\bibitem[An et~al.(2023)An, Zhang, Yang, Gupta, Huang, Luo, and Yin]{an2023latent}
Jie An, Songyang Zhang, Harry Yang, Sonal Gupta, Jia-Bin Huang, Jiebo Luo, and Xi Yin.
\newblock Latent-shift: Latent diffusion with temporal shift for efficient text-to-video generation.
\newblock \emph{arXiv preprint arXiv:2304.08477}, 2023.

\bibitem[Arnab et~al.(2021)Arnab, Dehghani, Heigold, Sun, Lu{\v{c}}i{\'c}, and Schmid]{arnab2021vivit}
Anurag Arnab, Mostafa Dehghani, Georg Heigold, Chen Sun, Mario Lu{\v{c}}i{\'c}, and Cordelia Schmid.
\newblock Vivit: A video vision transformer.
\newblock In \emph{Proceedings of the IEEE/CVF international conference on computer vision}, pages 6836--6846, 2021.

\bibitem[Blei et~al.(2017)Blei, Kucukelbir, and McAuliffe]{blei2017variational}
David~M Blei, Alp Kucukelbir, and Jon~D McAuliffe.
\newblock Variational inference: A review for statisticians.
\newblock \emph{Journal of the American statistical Association}, 112\penalty0 (518):\penalty0 859--877, 2017.

\bibitem[Esser et~al.(2021)Esser, Rombach, and Ommer]{esser2021taming}
Patrick Esser, Robin Rombach, and Bjorn Ommer.
\newblock Taming transformers for high-resolution image synthesis.
\newblock In \emph{Proceedings of the IEEE/CVF conference on computer vision and pattern recognition}, pages 12873--12883, 2021.

\bibitem[Fan et~al.(2024)Fan, Watkins, Du, Liu, Ryu, Boutilier, Abbeel, Ghavamzadeh, Lee, and Lee]{fan2024reinforcement}
Ying Fan, Olivia Watkins, Yuqing Du, Hao Liu, Moonkyung Ryu, Craig Boutilier, Pieter Abbeel, Mohammad Ghavamzadeh, Kangwook Lee, and Kimin Lee.
\newblock Reinforcement learning for fine-tuning text-to-image diffusion models.
\newblock \emph{Advances in Neural Information Processing Systems}, 36, 2024.

\bibitem[Fu et~al.(2023)Fu, Yu, Zhang, Fu, Su, Wang, and Bell]{fu2023tell}
Tsu-Jui Fu, Licheng Yu, Ning Zhang, Cheng-Yang Fu, Jong-Chyi Su, William~Yang Wang, and Sean Bell.
\newblock Tell me what happened: Unifying text-guided video completion via multimodal masked video generation.
\newblock In \emph{Proceedings of the IEEE/CVF Conference on Computer Vision and Pattern Recognition}, pages 10681--10692, 2023.

\bibitem[Gao et~al.(2023)Gao, Schulman, and Hilton]{gao2023scaling}
Leo Gao, John Schulman, and Jacob Hilton.
\newblock Scaling laws for reward model overoptimization.
\newblock In \emph{International Conference on Machine Learning}, pages 10835--10866. PMLR, 2023.

\bibitem[Ge et~al.(2023)Ge, Nah, Liu, Poon, Tao, Catanzaro, Jacobs, Huang, Liu, and Balaji]{ge2023preserve}
Songwei Ge, Seungjun Nah, Guilin Liu, Tyler Poon, Andrew Tao, Bryan Catanzaro, David Jacobs, Jia-Bin Huang, Ming-Yu Liu, and Yogesh Balaji.
\newblock Preserve your own correlation: A noise prior for video diffusion models.
\newblock In \emph{Proceedings of the IEEE/CVF International Conference on Computer Vision}, pages 22930--22941, 2023.

\bibitem[Guo et~al.(2023)Guo, Yang, Rao, Liang, Wang, Qiao, Agrawala, Lin, and Dai]{guo2023animatediff}
Yuwei Guo, Ceyuan Yang, Anyi Rao, Zhengyang Liang, Yaohui Wang, Yu Qiao, Maneesh Agrawala, Dahua Lin, and Bo Dai.
\newblock Animatediff: Animate your personalized text-to-image diffusion models without specific tuning.
\newblock \emph{arXiv preprint arXiv:2307.04725}, 2023.

\bibitem[He et~al.(2022)He, Yang, Zhang, Shan, and Chen]{he2022latent}
Yingqing He, Tianyu Yang, Yong Zhang, Ying Shan, and Qifeng Chen.
\newblock Latent video diffusion models for high-fidelity long video generation.
\newblock \emph{arXiv preprint arXiv:2211.13221}, 2022.

\bibitem[Ho et~al.(2022)Ho, Chan, Saharia, Whang, Gao, Gritsenko, Kingma, Poole, Norouzi, Fleet, et~al.]{ho2022imagen}
Jonathan Ho, William Chan, Chitwan Saharia, Jay Whang, Ruiqi Gao, Alexey Gritsenko, Diederik~P Kingma, Ben Poole, Mohammad Norouzi, David~J Fleet, et~al.
\newblock Imagen video: High definition video generation with diffusion models.
\newblock \emph{arXiv preprint arXiv:2210.02303}, 2022.

\bibitem[Hoffman et~al.(2013)Hoffman, Blei, Wang, and Paisley]{hoffman2013stochastic}
Matthew~D Hoffman, David~M Blei, Chong Wang, and John Paisley.
\newblock Stochastic variational inference.
\newblock \emph{Journal of Machine Learning Research}, 2013.

\bibitem[Kingma(2013)]{kingma2013auto}
Diederik~P Kingma.
\newblock Auto-encoding variational bayes.
\newblock \emph{arXiv preprint arXiv:1312.6114}, 2013.

\bibitem[Kirkpatrick et~al.(2017)Kirkpatrick, Pascanu, Rabinowitz, Veness, Desjardins, Rusu, Milan, Quan, Ramalho, Grabska-Barwinska, et~al.]{kirkpatrick2017overcoming}
James Kirkpatrick, Razvan Pascanu, Neil Rabinowitz, Joel Veness, Guillaume Desjardins, Andrei~A Rusu, Kieran Milan, John Quan, Tiago Ramalho, Agnieszka Grabska-Barwinska, et~al.
\newblock Overcoming catastrophic forgetting in neural networks.
\newblock \emph{Proceedings of the national academy of sciences}, 114\penalty0 (13):\penalty0 3521--3526, 2017.

\bibitem[Koley et~al.(2024)Koley, Bhunia, Sekhri, Sain, Chowdhury, Xiang, and Song]{koley2024s}
Subhadeep Koley, Ayan~Kumar Bhunia, Deeptanshu Sekhri, Aneeshan Sain, Pinaki~Nath Chowdhury, Tao Xiang, and Yi-Zhe Song.
\newblock It's all about your sketch: Democratising sketch control in diffusion models.
\newblock In \emph{Proceedings of the IEEE/CVF Conference on Computer Vision and Pattern Recognition}, pages 7204--7214, 2024.

\bibitem[Lee et~al.(2023)Lee, Liu, Ryu, Watkins, Du, Boutilier, Abbeel, Ghavamzadeh, and Gu]{lee2023aligning}
Kimin Lee, Hao Liu, Moonkyung Ryu, Olivia Watkins, Yuqing Du, Craig Boutilier, Pieter Abbeel, Mohammad Ghavamzadeh, and Shixiang~Shane Gu.
\newblock Aligning text-to-image models using human feedback.
\newblock \emph{arXiv preprint arXiv:2302.12192}, 2023.

\bibitem[Li et~al.(2018)Li, Min, Shen, Carlson, and Carin]{li2018video}
Yitong Li, Martin Min, Dinghan Shen, David Carlson, and Lawrence Carin.
\newblock Video generation from text.
\newblock In \emph{Proceedings of the AAAI conference on artificial intelligence}, 2018.

\bibitem[Lillicrap(2015)]{lillicrap2015continuous}
TP Lillicrap.
\newblock Continuous control with deep reinforcement learning.
\newblock \emph{arXiv preprint arXiv:1509.02971}, 2015.

\bibitem[Luo et~al.(2023)Luo, Chen, Zhang, Huang, Wang, Shen, Zhao, Zhou, and Tan]{luo2023videofusion}
Zhengxiong Luo, Dayou Chen, Yingya Zhang, Yan Huang, Liang Wang, Yujun Shen, Deli Zhao, Jingren Zhou, and Tieniu Tan.
\newblock Videofusion: Decomposed diffusion models for high-quality video generation.
\newblock \emph{arXiv preprint arXiv:2303.08320}, 2023.

\bibitem[Menapace et~al.(2024)Menapace, Siarohin, Skorokhodov, Deyneka, Chen, Kag, Fang, Stoliar, Ricci, Ren, et~al.]{menapace2024snap}
Willi Menapace, Aliaksandr Siarohin, Ivan Skorokhodov, Ekaterina Deyneka, Tsai-Shien Chen, Anil Kag, Yuwei Fang, Aleksei Stoliar, Elisa Ricci, Jian Ren, et~al.
\newblock Snap video: Scaled spatiotemporal transformers for text-to-video synthesis.
\newblock In \emph{Proceedings of the IEEE/CVF Conference on Computer Vision and Pattern Recognition}, pages 7038--7048, 2024.

\bibitem[Nichol et~al.(2021)Nichol, Dhariwal, Ramesh, Shyam, Mishkin, McGrew, Sutskever, and Chen]{nichol2021glide}
Alex Nichol, Prafulla Dhariwal, Aditya Ramesh, Pranav Shyam, Pamela Mishkin, Bob McGrew, Ilya Sutskever, and Mark Chen.
\newblock Glide: Towards photorealistic image generation and editing with text-guided diffusion models.
\newblock \emph{arXiv preprint arXiv:2112.10741}, 2021.

\bibitem[Podell et~al.(2023)Podell, English, Lacey, Blattmann, Dockhorn, M{\"u}ller, Penna, and Rombach]{podell2023sdxl}
Dustin Podell, Zion English, Kyle Lacey, Andreas Blattmann, Tim Dockhorn, Jonas M{\"u}ller, Joe Penna, and Robin Rombach.
\newblock Sdxl: Improving latent diffusion models for high-resolution image synthesis.
\newblock \emph{arXiv preprint arXiv:2307.01952}, 2023.

\bibitem[Qing et~al.(2024)Qing, Zhang, Wang, Wang, Wei, Zhang, Gao, and Sang]{qing2024hierarchical}
Zhiwu Qing, Shiwei Zhang, Jiayu Wang, Xiang Wang, Yujie Wei, Yingya Zhang, Changxin Gao, and Nong Sang.
\newblock Hierarchical spatio-temporal decoupling for text-to-video generation.
\newblock In \emph{Proceedings of the IEEE/CVF Conference on Computer Vision and Pattern Recognition}, pages 6635--6645, 2024.

\bibitem[Radford et~al.(2021)Radford, Kim, Hallacy, Ramesh, Goh, Agarwal, Sastry, Askell, Mishkin, Clark, et~al.]{radford2021learning}
Alec Radford, Jong~Wook Kim, Chris Hallacy, Aditya Ramesh, Gabriel Goh, Sandhini Agarwal, Girish Sastry, Amanda Askell, Pamela Mishkin, Jack Clark, et~al.
\newblock Learning transferable visual models from natural language supervision.
\newblock In \emph{International conference on machine learning}, pages 8748--8763. PMLR, 2021.

\bibitem[Rafailov et~al.(2024)Rafailov, Sharma, Mitchell, Manning, Ermon, and Finn]{rafailov2024direct}
Rafael Rafailov, Archit Sharma, Eric Mitchell, Christopher~D Manning, Stefano Ermon, and Chelsea Finn.
\newblock Direct preference optimization: Your language model is secretly a reward model.
\newblock \emph{Advances in Neural Information Processing Systems}, 36, 2024.

\bibitem[Ramesh et~al.(2021)Ramesh, Pavlov, Goh, Gray, Voss, Radford, Chen, and Sutskever]{ramesh2021zero}
Aditya Ramesh, Mikhail Pavlov, Gabriel Goh, Scott Gray, Chelsea Voss, Alec Radford, Mark Chen, and Ilya Sutskever.
\newblock Zero-shot text-to-image generation.
\newblock In \emph{International conference on machine learning}, pages 8821--8831. Pmlr, 2021.

\bibitem[Ramesh et~al.(2022)Ramesh, Dhariwal, Nichol, Chu, and Chen]{ramesh2022hierarchical}
Aditya Ramesh, Prafulla Dhariwal, Alex Nichol, Casey Chu, and Mark Chen.
\newblock Hierarchical text-conditional image generation with clip latents.
\newblock \emph{arXiv preprint arXiv:2204.06125}, 1\penalty0 (2):\penalty0 3, 2022.

\bibitem[Rombach et~al.(2022)Rombach, Blattmann, Lorenz, Esser, and Ommer]{rombach2022high}
Robin Rombach, Andreas Blattmann, Dominik Lorenz, Patrick Esser, and Bj{\"o}rn Ommer.
\newblock High-resolution image synthesis with latent diffusion models.
\newblock In \emph{Proceedings of the IEEE/CVF conference on computer vision and pattern recognition}, pages 10684--10695, 2022.

\bibitem[Saharia et~al.(2022)Saharia, Chan, Saxena, Li, Whang, Denton, Ghasemipour, Gontijo~Lopes, Karagol~Ayan, Salimans, et~al.]{saharia2022photorealistic}
Chitwan Saharia, William Chan, Saurabh Saxena, Lala Li, Jay Whang, Emily~L Denton, Kamyar Ghasemipour, Raphael Gontijo~Lopes, Burcu Karagol~Ayan, Tim Salimans, et~al.
\newblock Photorealistic text-to-image diffusion models with deep language understanding.
\newblock \emph{Advances in neural information processing systems}, 35:\penalty0 36479--36494, 2022.

\bibitem[Saito et~al.(2017)Saito, Matsumoto, and Saito]{saito2017temporal}
Masaki Saito, Eiichi Matsumoto, and Shunta Saito.
\newblock Temporal generative adversarial nets with singular value clipping.
\newblock In \emph{Proceedings of the IEEE international conference on computer vision}, pages 2830--2839, 2017.

\bibitem[Schulman et~al.(2017)Schulman, Wolski, Dhariwal, Radford, and Klimov]{schulman2017proximal}
John Schulman, Filip Wolski, Prafulla Dhariwal, Alec Radford, and Oleg Klimov.
\newblock Proximal policy optimization algorithms.
\newblock \emph{arXiv preprint arXiv:1707.06347}, 2017.

\bibitem[Shi et~al.(2024)Shi, Xiong, Lin, and Jung]{shi2024instantbooth}
Jing Shi, Wei Xiong, Zhe Lin, and Hyun~Joon Jung.
\newblock Instantbooth: Personalized text-to-image generation without test-time finetuning.
\newblock In \emph{Proceedings of the IEEE/CVF Conference on Computer Vision and Pattern Recognition}, pages 8543--8552, 2024.

\bibitem[Sohl-Dickstein et~al.(2015)Sohl-Dickstein, Weiss, Maheswaranathan, and Ganguli]{sohl2015deep}
Jascha Sohl-Dickstein, Eric Weiss, Niru Maheswaranathan, and Surya Ganguli.
\newblock Deep unsupervised learning using nonequilibrium thermodynamics.
\newblock In \emph{International conference on machine learning}, pages 2256--2265. PMLR, 2015.

\bibitem[Song et~al.(2020)Song, Meng, and Ermon]{song2020denoising}
Jiaming Song, Chenlin Meng, and Stefano Ermon.
\newblock Denoising diffusion implicit models.
\newblock \emph{arXiv preprint arXiv:2010.02502}, 2020.

\bibitem[Stiennon et~al.(2020)Stiennon, Ouyang, Wu, Ziegler, Lowe, Voss, Radford, Amodei, and Christiano]{stiennon2020learning}
Nisan Stiennon, Long Ouyang, Jeffrey Wu, Daniel Ziegler, Ryan Lowe, Chelsea Voss, Alec Radford, Dario Amodei, and Paul~F Christiano.
\newblock Learning to summarize with human feedback.
\newblock \emph{Advances in Neural Information Processing Systems}, 33:\penalty0 3008--3021, 2020.

\bibitem[Tulyakov et~al.(2018)Tulyakov, Liu, Yang, and Kautz]{tulyakov2018mocogan}
Sergey Tulyakov, Ming-Yu Liu, Xiaodong Yang, and Jan Kautz.
\newblock Mocogan: Decomposing motion and content for video generation.
\newblock In \emph{Proceedings of the IEEE conference on computer vision and pattern recognition}, pages 1526--1535, 2018.

\bibitem[Wallace et~al.(2024)Wallace, Dang, Rafailov, Zhou, Lou, Purushwalkam, Ermon, Xiong, Joty, and Naik]{wallace2024diffusion}
Bram Wallace, Meihua Dang, Rafael Rafailov, Linqi Zhou, Aaron Lou, Senthil Purushwalkam, Stefano Ermon, Caiming Xiong, Shafiq Joty, and Nikhil Naik.
\newblock Diffusion model alignment using direct preference optimization.
\newblock In \emph{Proceedings of the IEEE/CVF Conference on Computer Vision and Pattern Recognition}, pages 8228--8238, 2024.

\bibitem[Wang et~al.(2023)Wang, Chen, Qiu, Chen, Wu, Lin, He, and Liu]{wang2023crossformer++}
Wenxiao Wang, Wei Chen, Qibo Qiu, Long Chen, Boxi Wu, Binbin Lin, Xiaofei He, and Wei Liu.
\newblock Crossformer++: A versatile vision transformer hinging on cross-scale attention.
\newblock \emph{IEEE Transactions on Pattern Analysis and Machine Intelligence}, 2023.

\bibitem[Wang et~al.(2004)Wang, Bovik, Sheikh, and Simoncelli]{wang2004image}
Zhou Wang, Alan~C Bovik, Hamid~R Sheikh, and Eero~P Simoncelli.
\newblock Image quality assessment: from error visibility to structural similarity.
\newblock \emph{IEEE transactions on image processing}, 13\penalty0 (4):\penalty0 600--612, 2004.

\bibitem[Wu et~al.(2021)Wu, Huang, Zhang, Li, Ji, Yang, Sapiro, and Duan]{wu2021godiva}
Chenfei Wu, Lun Huang, Qianxi Zhang, Binyang Li, Lei Ji, Fan Yang, Guillermo Sapiro, and Nan Duan.
\newblock Godiva: Generating open-domain videos from natural descriptions.
\newblock \emph{arXiv preprint arXiv:2104.14806}, 2021.

\bibitem[Wu et~al.(2022)Wu, Liang, Ji, Yang, Fang, Jiang, and Duan]{wu2022nuwa}
Chenfei Wu, Jian Liang, Lei Ji, Fan Yang, Yuejian Fang, Daxin Jiang, and Nan Duan.
\newblock N{\"u}wa: Visual synthesis pre-training for neural visual world creation.
\newblock In \emph{European conference on computer vision}, pages 720--736. Springer, 2022.

\bibitem[Wu et~al.(2023)Wu, Chen, Yang, Guo, Li, and Zhang]{wu2023lamp}
Ruiqi Wu, Liangyu Chen, Tong Yang, Chunle Guo, Chongyi Li, and Xiangyu Zhang.
\newblock Lamp: Learn a motion pattern for few-shot-based video generation.
\newblock \emph{arXiv preprint arXiv:2310.10769}, 2023.

\bibitem[Xing et~al.(2024)Xing, Dai, Hu, Wu, and Jiang]{xing2024simda}
Zhen Xing, Qi Dai, Han Hu, Zuxuan Wu, and Yu-Gang Jiang.
\newblock Simda: Simple diffusion adapter for efficient video generation.
\newblock In \emph{Proceedings of the IEEE/CVF Conference on Computer Vision and Pattern Recognition}, pages 7827--7839, 2024.

\bibitem[Yan et~al.(2021)Yan, Zhang, Abbeel, and Srinivas]{yan2021videogpt}
Wilson Yan, Yunzhi Zhang, Pieter Abbeel, and Aravind Srinivas.
\newblock Videogpt: Video generation using vq-vae and transformers.
\newblock \emph{arXiv preprint arXiv:2104.10157}, 2021.

\bibitem[Yang et~al.(2024)Yang, Tao, Lyu, Ge, Chen, Shen, Zhu, and Li]{yang2024using}
Kai Yang, Jian Tao, Jiafei Lyu, Chunjiang Ge, Jiaxin Chen, Weihan Shen, Xiaolong Zhu, and Xiu Li.
\newblock Using human feedback to fine-tune diffusion models without any reward model.
\newblock In \emph{Proceedings of the IEEE/CVF Conference on Computer Vision and Pattern Recognition}, pages 8941--8951, 2024.

\bibitem[Yu et~al.(2022{\natexlab{a}})Yu, Xu, Koh, Luong, Baid, Wang, Vasudevan, Ku, Yang, Ayan, et~al.]{yu2022scaling}
Jiahui Yu, Yuanzhong Xu, Jing~Yu Koh, Thang Luong, Gunjan Baid, Zirui Wang, Vijay Vasudevan, Alexander Ku, Yinfei Yang, Burcu~Karagol Ayan, et~al.
\newblock Scaling autoregressive models for content-rich text-to-image generation.
\newblock \emph{arXiv preprint arXiv:2206.10789}, 2\penalty0 (3):\penalty0 5, 2022{\natexlab{a}}.

\bibitem[Yu et~al.(2022{\natexlab{b}})Yu, Tack, Mo, Kim, Kim, Ha, and Shin]{yu2022generating}
Sihyun Yu, Jihoon Tack, Sangwoo Mo, Hyunsu Kim, Junho Kim, Jung-Woo Ha, and Jinwoo Shin.
\newblock Generating videos with dynamics-aware implicit generative adversarial networks.
\newblock \emph{arXiv preprint arXiv:2202.10571}, 2022{\natexlab{b}}.

\bibitem[Zhang et~al.(2022)Zhang, Tao, and Chen]{zhang2022gddim}
Qinsheng Zhang, Molei Tao, and Yongxin Chen.
\newblock gddim: Generalized denoising diffusion implicit models.
\newblock \emph{arXiv preprint arXiv:2206.05564}, 2022.

\end{thebibliography}
}

\clearpage

\setcounter{page}{1}
\appendix
\onecolumn 
\begin{center}
\Large
\textbf{HuViDPO: Enhancing Video Generation through Direct Preference
Optimization for Human-Centric Alignment} \vspace{0.5em} Supplementary Material 
\end{center}

\section{Formula Derivation for DPO}
This section provides a structured derivation of the Direct Preference Optimization (DPO) formula. First, the technical background necessary to understand the DPO approach is presented, focusing on the concepts and preliminary models used. Next, the objective function of DPO is detailed, deriving it based on the reward function and KL divergence to establish a preference-based optimization framework.

\subsection{Technical Background}
The core idea of Direct Preference Optimization (DPO) is to utilize preference data directly for policy optimization, avoiding the dependency on reward model-based methods. Given a prompt \( x \), two responses \( (y_1, y_2) \sim \Pi_{\text{SFT}}(y | x) \) are generated. By manually annotating, the preferences of \( y_1 \) and \( y_2 \) are compared, resulting in a preference outcome \( y_w \succ y_l | x \), where \( w \) and \( l \) denote the "win" and "lose" responses, respectively.

To formalize preference probabilities, a reward model \( r \) is introduced such that for two generated responses \( y_1 \) and \( y_2 \), the probability of \( y_1 \) being preferred over \( y_2 \) is expressed as:

\begin{equation}
p(y_1 > y_2) = \frac{r^*(x, y_1)}{r^*(x, y_1) + r^*(x, y_2)}.
\end{equation}

Here, the reward function \( r^*(x, y) \) quantifies the quality of response \( y \) relative to prompt \( x \), and the probability \( p(y_1 > y_2) \) reflects the likelihood of choosing \( y_1 \) over \( y_2 \) based on this reward.

To ensure positive values in the reward model, the Bradley-Terry model is applied, representing the preference probability \( p(y_w \succ y_l | x) \) as:

\begin{equation}
p^*(y_w \succ y_l | x) = \frac{\exp(r^*(x, y_1))}{\exp(r^*(x, y_1)) + \exp(r^*(x, y_2))}.
\end{equation}

This exponential form stabilizes the reward values and maintains positive preference probabilities.

\subsection{DPO Objective Function}
The goal of DPO is to maximize rewards while aligning the policy with a baseline model. The objective function is defined as:

\begin{equation}
\max_\pi \mathbb{E}_{x \in X, y \in \pi} \left[ r(x, y) \right] - \beta \cdot \mathbb{D}_{\text{KL}} \left[ \pi(y | x) || \pi_{\text{ref}}(y | x) \right],
\end{equation}

where \( D_{\text{KL}} \) denotes the Kullback-Leibler divergence between the learned policy \( \pi \) and a reference policy \( \pi_{\text{ref}} \), enforcing consistency with the baseline model. To simplify, this objective is reformulated as a minimization problem:

\begin{equation}
\min_\pi \mathbb{E}_{x \in X, y \in \pi} \left[ \log \frac{\pi(y | x)}{\pi^*(y | x)} - \log Z(x) \right],
\end{equation}

where \( Z(x) \) is defined as:

\begin{equation}
Z(x) = \sum_y \pi_{\text{ref}}(y | x) \exp \left( \frac{1}{\beta} r(x, y) \right).
\end{equation}

This reformulation leads to the final optimization objective:

\begin{equation}
\min_\pi \mathbb{E}_{x \sim D} \left[ \mathbb{D}_{\text{KL}}(\pi(y | x) || \pi^*(y | x)) \right].
\end{equation}

Under the minimization of KL divergence, the policy \( \pi(y | x) \) adheres to the following form:

\begin{equation}
\pi(y | x) = \pi^*(y | x) = \frac{1}{Z(x)} \pi_{\text{ref}}(y | x) \cdot \exp \left( \frac{1}{\beta} r(x, y) \right).
\end{equation}

Reversing this equation yields the reward function:

\begin{equation}
r^*(x, y) = \beta \log \frac{\pi(y \mid x)}{\pi_{\text{ref}}(y \mid x)}.
\end{equation}

Incorporating the Bradley-Terry model, the cross-entropy loss function \(\mathcal{L}\) is defined, which quantifies the difference between the preferred and non-preferred responses. This loss function is essential for deriving the gradient necessary to optimize the DPO objective:

\begin{equation}
\mathcal{L} = -\mathbb{E}_{(x, y_w, y_l) \sim D} \left[ \ln \sigma \left( \beta \log \frac{\pi(y_w | x)}{\pi_{\text{ref}}(y_w | x)} - \beta \log \frac{\pi(y_l | x)}{\pi_{\text{ref}}(y_l | x)} \right) \right],
\end{equation}

where \( \sigma \) denotes the sigmoid function, which maps the difference in log-probabilities to a range of \([0, 1]\). Differentiating \(\mathcal{L}\) provides the gradient needed to optimize the DPO objective with respect to the preference data.

\section{Applying DPO Strategy into T2I tasks}

In adapting Dynamic Preference Optimization (DPO) to text-to-image tasks (T2I tasks), we consider a setting with a fixed dataset \( D = \{(c, x_0^w, x_0^l)\} \). In this dataset, each example contains a prompt \( c \) and a pair of images \( (x_0^w, x_0^l) \) generated by a reference model \( p_{\text{ref}} \), where \( x_0^w \succ x_0^l \) indicates that humans prefer \( x_0^w \) over \( x_0^l \). The goal is to train a new model \( p_\theta \) so that its generated images align with human preferences, rather than merely imitating the reference model. However, directly computing the distribution \( p_\theta(x_0 | c) \) is highly complex, as it requires marginalizing over all possible generation paths \( (x_1, \ldots, x_T) \) to produce \( x_0 \), which is practically infeasible.

To address this challenge, researchers leverage Evidence Lower Bound (ELBO) by introducing latent variables \( x_{1:T} \). The reward function \( R(c, x_{0:T}) \) is defined to measure the quality of the entire generation path, allowing the expected reward \( r(c, x_0) \) for given \( c \) and \( x_0 \) to be formulated as:

\begin{equation}
r(c, x_0) = \mathbb{E}_{p_\theta(x_{1:T} | x_0, c)} [R(c, x_{0:T})]
\end{equation}

In DPO, a KL regularization term is also included to constrain the generated distribution relative to the reference distribution. Here, an upper bound on the KL divergence is used, converting it to a joint KL divergence:

\begin{equation}
\mathbb{D}_{KL}[p_\theta(x_{0:T} | c) \parallel p_{\text{ref}}(x_{0:T} | c)]
\end{equation}

This upper bound ensures that the distribution of the generated model \( p_\theta(x_{0:T} | c) \) remains consistent with the reference model \( p_{\text{ref}}(x_{0:T} | c) \), preserving the model’s generation capabilities while optimizing human preference alignment. Plugging in this KL divergence upper bound and the reward function \( r(c, x_0) \) into the objective function, we obtain:

\begin{equation}
\max_{p_\theta} \mathbb{E}_{c \sim \mathcal{D}_c, x_{0:T} \sim p_\theta(x_{0:T} | c)} [r(c, x_0)] - \beta \mathbb{D}_{KL} [p_\theta(x_{0:T} | c) \parallel p_{\text{ref}}(x_{0:T} | c)]
\end{equation}

This objective function resembles the structure of Equation (5) but is defined over the path \( x_{0:T} \). Its primary goal is to maximize the reward for the reverse process \( p_\theta(x_{0:T}) \) while maintaining distributional consistency with the original reference reverse process. To optimize this objective, the conditional distribution \( p_\theta(x_{0:T}) \) is directly used. The final loss function \( L_{\text{DPO-T2I}}(\theta) \) is expressed as follows:

\begin{equation}
\mathcal{L}_{\text{Image}}(\theta) = -\mathbb{E}_{(x_0^w, x_0^l) \sim D} \log \sigma \left( \beta \mathbb{E}_{x_{1:T}^w \sim p_\theta(x_{1:T} | x_0^w), \, x_{1:T}^l \sim p_\theta(x_{1:T} | x_0^l)} \left[ \log \frac{p_\theta(x_{0:T}^w)}{p_{\text{ref}}(x_{0:T}^w)} - \log \frac{p_\theta(x_{0:T}^l)}{p_{\text{ref}}(x_{0:T}^l)} \right] \right)
\end{equation}

By applying Jensen's inequality, the expectation can be moved outside of the \(\log \sigma\) function, resulting in an upper bound. This simplifies the formula and can facilitate optimization. After applying Jensen's inequality, the loss function's upper bound is given by:

\begin{equation}
\mathcal{L}_{\text{Image}}(\theta) \leq -\mathbb{E}_{(x_0^w, x_0^l) \sim D} \, \mathbb{E}_{x_{1:T}^w \sim p_\theta(x_{1:T} | x_0^w), \, x_{1:T}^l \sim p_\theta(x_{1:T} | x_0^l)} \, \log \sigma \left( \beta \left[ \log \frac{p_\theta(x_{0:T}^w)}{p_{\text{ref}}(x_{0:T}^w)} - \log \frac{p_\theta(x_{0:T}^l)}{p_{\text{ref}}(x_{0:T}^l)} \right] \right)
\end{equation}

The final loss function maximizes the reward \( r(c, x_0) \) aligned with human preference while ensuring that \( p_\theta \) stays consistent with the reference distribution \( p_{\text{ref}} \) through the KL regularization term. This effectively aligns human preferences in the T2I task's DPO process, enabling the model not only to generate high-quality images but also to reflect human visual preferences and intuitive aesthetics.

\section{Details of User Study}

\subsection{Objective}
To evaluate the performance of the proposed T2V method, we conduct this user study and compare the generated results with three classic open-source T2V methods: LVDM, AnimateDiff, and LAMP. The evaluation focuses on three key criteria: Video-Text Alignment, Visual Quality, and Frame Consistency.

\subsection{Methodology}

\textbf{Video Generation.} We select a set of 14 prompts to generate videos, including dynamic content featuring fireworks, waterfall, smile, horse, guitar, birds flying, helicopter, and rain, covering a total of 8 types. For each prompt, videos are produced using four models. This process resulted in a total of 56 videos.

\vspace{0.1cm}
\noindent\textbf{Study Procedure.} Each participant is asked to evaluate a series of videos. For each video, participants are instructed to provide three separate scores (each ranging from 1 to 5, with 1 being the lowest and 5 being the highest) based on the following criteria:

\begin{itemize}
    \item \textbf{Video-Text Alignment:} Does the video accurately reflect the description?
    \begin{itemize}
        \item  1 point: The video does not align with the description; it is almost impossible to recognize elements from the text in the video.
        \item  3 points: The video partially aligns with the description, showing some relevant content, but with noticeable mismatches or missing details.
        \item  5 points: The video perfectly aligns with the description, accurately reflecting all key elements of the text.
    \end{itemize}

    \item \textbf{Visual Quality:} How clear and visually appealing is the video?
    \begin{itemize}
        \item  1 point: The video is blurry, with severe pixelation or visual defects.
        \item  3 points: The video quality is acceptable; the main content is recognizable, though some details are rough.
        \item  5 points: The video quality is excellent, with high clarity and well-represented details.
    \end{itemize}

    \item \textbf{Frame Consistency:} How smooth and seamless are the transitions between frames?
    \begin{itemize}
        \item  1 point: Frames are highly inconsistent, appearing jumpy or choppy, disrupting the viewing experience.
        \item  3 points: Frames are somewhat consistent, but transitions have some roughness or slight jumps.
        \item  5 points: Frames are consistently smooth, delivering a natural and seamless viewing experience.
    \end{itemize}
\end{itemize}

Participants are instructed to score each video independently and provide honest feedback.

\textbf{Data Collection.} We collect test data using an online form. Scores for each criterion are averaged across participants and videos. Statistical analysis is then conducted based on the data.

\subsection{Results Interpretation}
The results are shown in ~\cref{tab:example2}. Due to file size limitations for submissions, we are temporarily unable to publicly share the specific video examples and their scoring details involved in the user study.

\section{Experiment Result}

This section demonstrates the powerful generative capabilities of the HuViDPO method, which can produce high-quality, human-preferred videos across various visual styles, including pixel art, Claymation, cartoon, and realistic styles. \textcolor[HTML]{94CB6B}{\textit{\textbf{Green}}} highlights indicate the subject category, \textcolor[HTML]{18D8FA}{\textit{\textbf{blue}}} highlights represent motion-related information, and \textcolor[HTML]{E3954F}{\textit{\textbf{orange}}} highlights denote the background.

\begin{figure*}[htbp]
    \centering
    \includegraphics[width=1\textwidth]{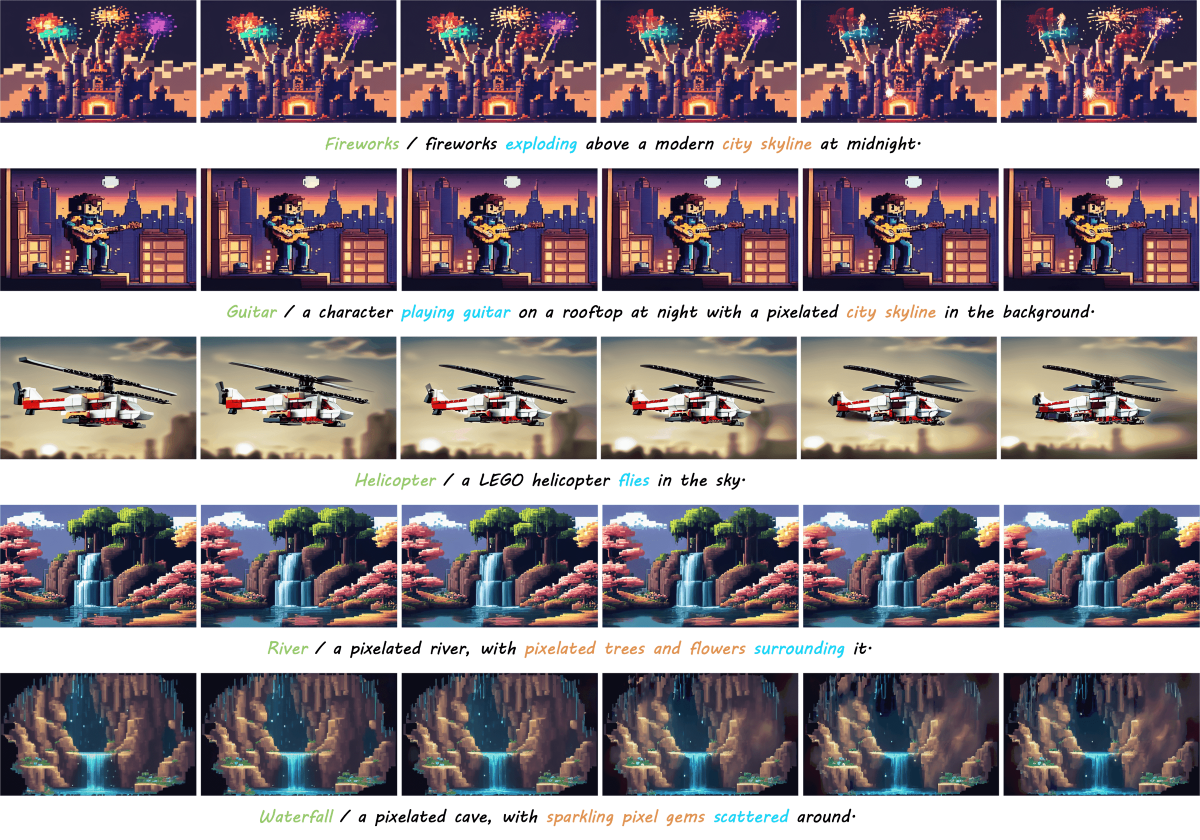}
    \caption{Pixel-art-style video generated by our method.}
\end{figure*}

\begin{figure*}[htbp]
    \centering
    \includegraphics[width=1\textwidth]{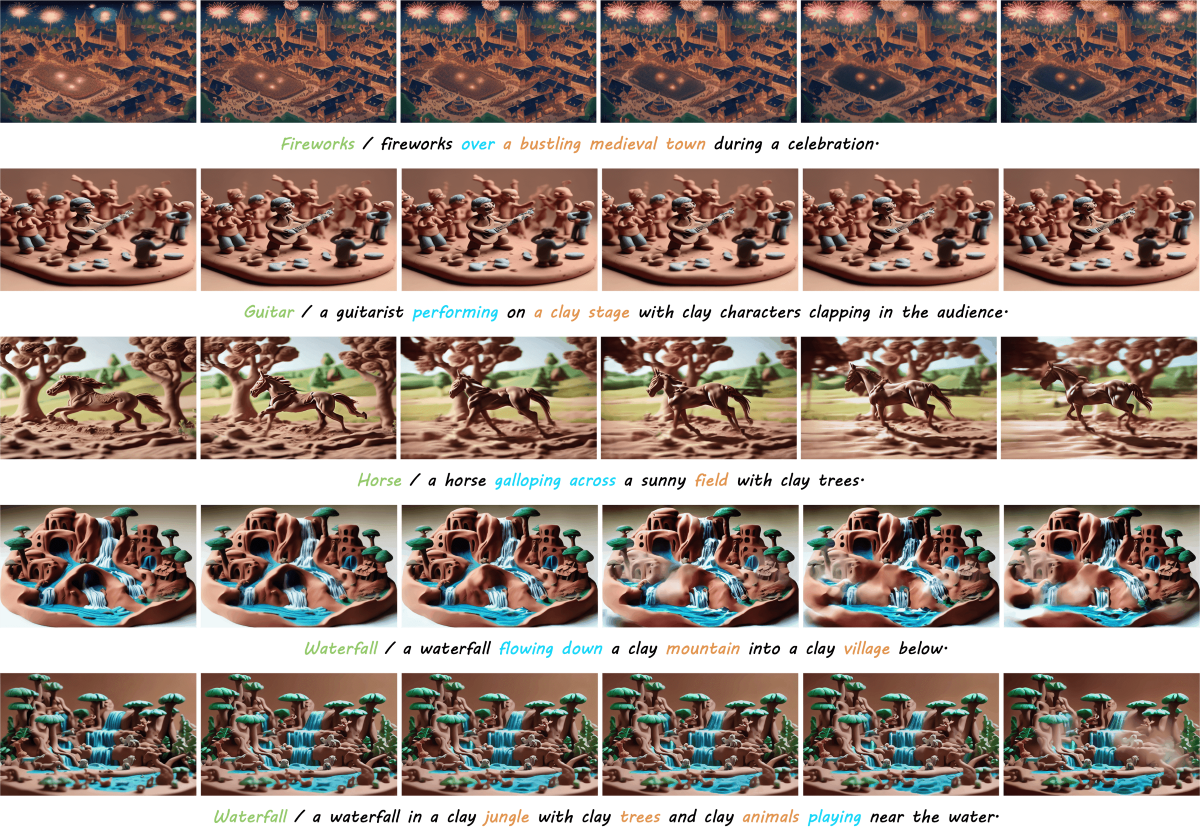}
    \caption{Claymation-style video generated by our method.}
\end{figure*}

\begin{figure*}[htbp]
    \centering
    \includegraphics[width=1\textwidth]{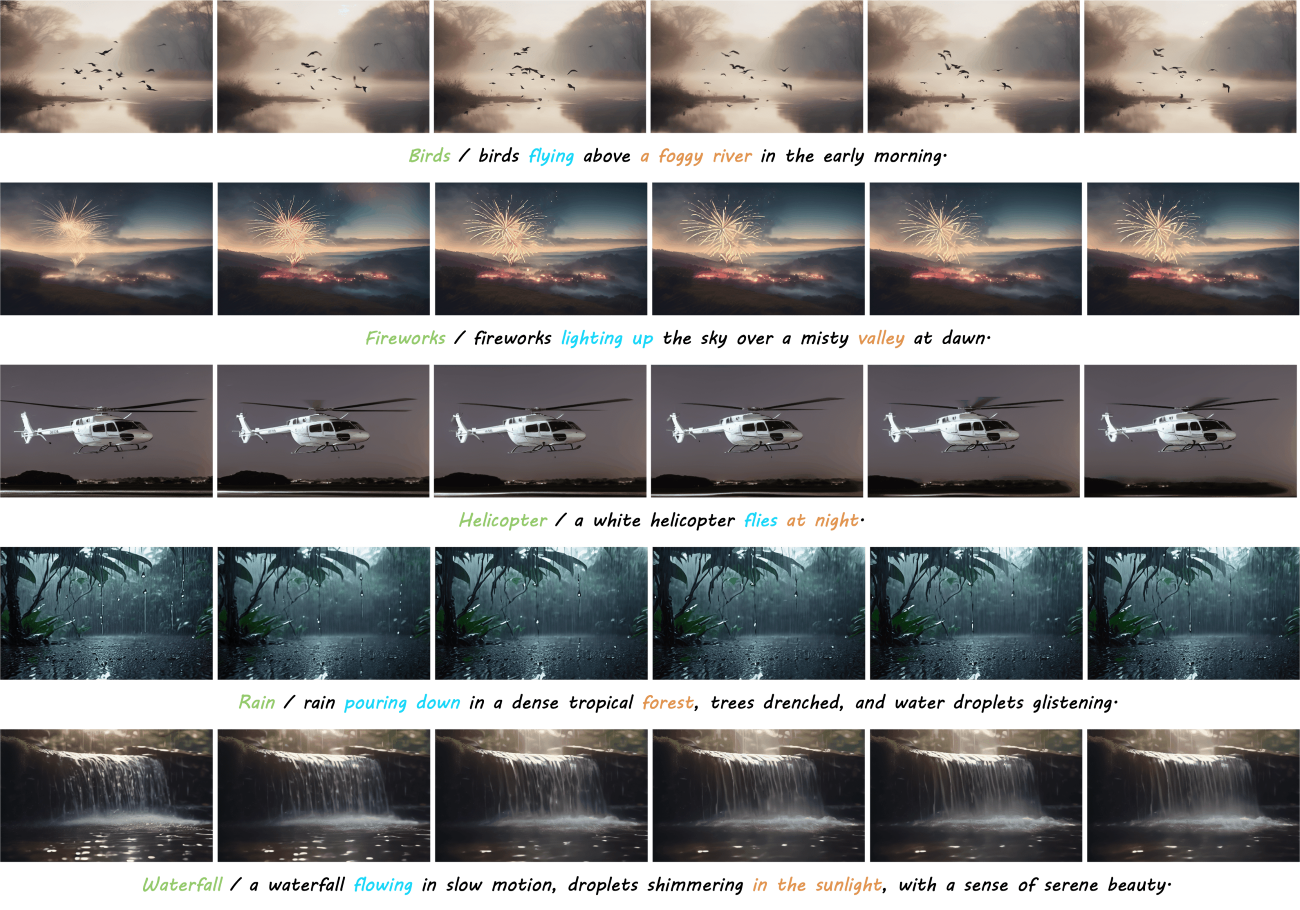}
    \caption{Realistic-style video generated by our method.}
\end{figure*}

\begin{figure*}[htbp]
    \centering
    \includegraphics[width=1\textwidth]{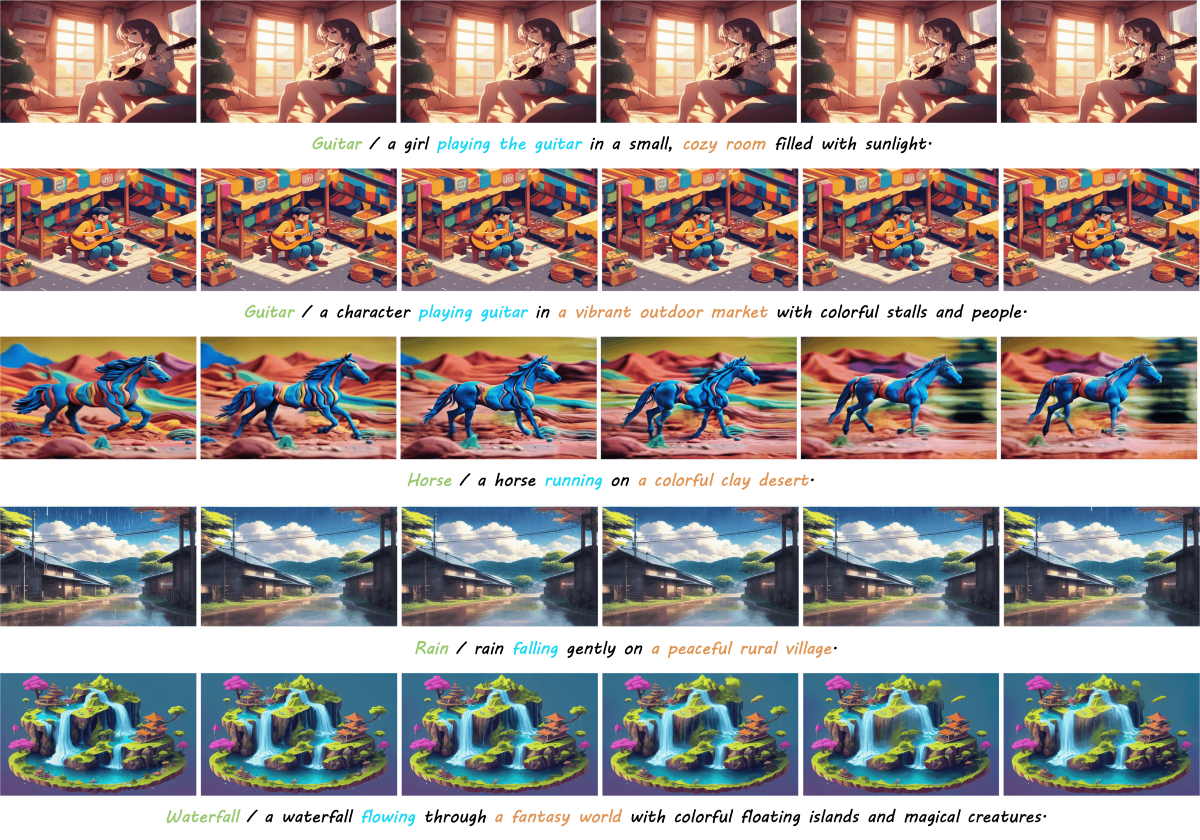}
    \caption{Cartoon-style video generated by our method.}
\end{figure*}


\end{document}